\documentclass{article} 
\usepackage{iclr2024_conference,times}

\usepackage{amsmath,amsfonts,bm}

\def\eqref#1{equation~\ref{#1}}

\def\1{\bm{1}}

\DeclareMathAlphabet{\mathsfit}{\encodingdefault}{\sfdefault}{m}{sl}
\SetMathAlphabet{\mathsfit}{bold}{\encodingdefault}{\sfdefault}{bx}{n}

\usepackage{hyperref}
\usepackage{url}
\usepackage{pifont}
\usepackage{longtable}
\usepackage{booktabs}
\usepackage{forest}
\usepackage{xcolor}

\usepackage{geometry}
\usepackage{tikz}
\usetikzlibrary{trees,shapes.geometric,calc}

\usepackage{subcaption}  

\title{Multilingual Prompt Engineering in Large Language Models: A Survey Across NLP Tasks}

\author{Shubham Vatsal \& Harsh Dubey  \\
Department of Computer Science\\
New York University, CIMS\\
New York, USA \\
\texttt{\{sv2128,hd2225\}@nyu.edu}
\And
  Aditi Singh \\
  Department of Computer Science\\
  Cleveland State University\\
  Cleveland, USA \\
  \texttt{a.singh22@csuohio.edu}
}

\iclrfinalcopy 

\begin{document}

\maketitle

\begin{abstract}
Large language models (LLMs) have demonstrated impressive performance across a wide range of Natural Language Processing (NLP) tasks. However, ensuring their effectiveness across multiple languages presents unique challenges. Multilingual prompt engineering has emerged as a key approach to enhance LLMs’ capabilities in diverse linguistic settings without requiring extensive parameter re-training or fine-tuning. With growing interest in multilingual prompt engineering over the past two to three years, researchers have explored various strategies to improve LLMs’ performance across languages and NLP tasks. By crafting structured natural language prompts, researchers have successfully extracted knowledge from LLMs across different languages, making these techniques an accessible pathway for a broader audience, including those without deep expertise in machine learning, to harness the capabilities of LLMs. In this paper, we survey and categorize different multilingual prompting techniques based on the NLP tasks they address across a diverse set of datasets that collectively span around 250 languages. We further highlight the LLMs employed, present a taxonomy of approaches and discuss potential state-of-the-art (SoTA) methods for specific multilingual datasets. Additionally, we derive a range of insights across language families and resource levels (high-resource vs. low-resource), including analyses such as the distribution of NLP tasks by language resource type and the frequency of prompting methods across different language families. Our survey reviews 36 research papers covering 39 prompting techniques applied to 30 multilingual NLP tasks, with the majority of these studies published in the last two years.
\end{abstract}


\section{Introduction}


With the advent of LLMs, significant advancements have been made in NLP. These models are trained on vast corpora of text, containing millions or even billions of tokens. Research has shown that increasing the number of model parameters enhances the performance of machine learning models and LLMs have followed this trend. They have achieved unprecedented performance across a wide range of NLP tasks \cite{chang2023survey}, garnering substantial interest from academia and various industries, including medicine, law and finance. As LLMs become more capable, their deployment in real-world applications has accelerated, further highlighting their transformative potential. The current phase of LLM research emphasizes their reasoning capabilities through prompts, rather than merely predicting the next token. This shift has given rise to an emerging field of study known as prompt engineering. Researchers are now focused not only on scaling models, but also on optimizing their interaction with natural language instructions to unlock higher levels of understanding and task-specific performance.

Hard (Discrete) prompt engineering \footnote{From here on, we refer to hard (discrete) prompt engineering simply as prompt engineering unless differentiation from soft prompt engineering is necessary.} is the process of designing and crafting natural language instructions or prompts to guide LLMs in extracting knowledge and performing tasks in an organized and efficient manner. This approach sets itself apart from earlier conventional models by relying solely on the embedded knowledge within LLMs, eliminating the need for extensive parameter re-training or fine-tuning tailored to specific NLP tasks.

Traditional models typically require task-specific adjustments and training on large labeled datasets. In contrast, prompt engineering leverages the pre-existing knowledge that LLMs acquire during their initial training, allowing researchers to directly utilize the model’s capabilities through carefully crafted prompts. This eliminates the need for complex retraining, which is often both time-consuming and resource-intensive. Understanding the intricate details of the model's parameters in terms of real-world knowledge is beyond human comprehension due to the sheer scale of data that LLMs are trained on. This gap in understanding is one of the key reasons prompt engineering has gained significant attention. By utilizing natural language as a communication interface, prompt engineering facilitates seamless interaction between researchers and LLMs. It offers a powerful tool for achieving specific goals across a wide variety of NLP tasks, including question answering, text generation, reasoning and problem-solving. As a result, prompt engineering has become a critical area of research, enabling more efficient and accessible use of LLMs without requiring specialized expertise in machine learning model development.

While prompt engineering has significantly advanced the capabilities of LLMs, most early research in this domain has primarily focused on English language tasks. However, as LLMs continue to expand their applicability across diverse linguistic and cultural contexts, the need for multilingual prompt engineering has become increasingly evident. Multilingual prompt engineering involves designing prompts that effectively guide LLMs in processing, understanding and generating text across multiple languages. This is particularly crucial for NLP tasks that require cross-lingual generalization, machine translation, multilingual question answering and sentiment analysis. One of the major challenges in multilingual prompt engineering is ensuring consistent performance across languages, as LLMs often exhibit disparities in proficiency depending on the availability of training data for different languages. High-resource languages, such as English, Spanish and Chinese, typically yield stronger results compared to low-resource languages, which may suffer from limited linguistic representation in pre-training corpora. Researchers have explored various strategies to mitigate these disparities, including prompt adaptation, translation-based prompting and few-shot learning with cross-lingual examples. Furthermore, cultural and linguistic nuances play a significant role in the effectiveness of multilingual prompts. Direct translations of prompts do not always yield optimal results due to differences in syntactic structures, idiomatic expressions and pragmatic interpretations across languages. To address this, multilingual prompt engineering often incorporates techniques such as language-specific prompt tuning and hybrid approaches that leverage both native-language prompts and interlingual prompting strategies. Given the increasing global adoption of LLMs, multilingual prompt engineering is essential for ensuring equitable access to advanced NLP technologies across languages and cultures. Research in this field continues to evolve, exploring ways to enhance prompt effectiveness while addressing the linguistic diversity and challenges inherent in multilingual NLP tasks.


\subsection{Focus of Our Survey}

The rise of LLMs has led to significant advancements in NLP, yet ensuring their performance across multiple languages remains a challenge. Multilingual prompt engineering has emerged as a pivotal approach to enhance LLMs’ cross-linguistic capabilities without requiring extensive fine-tuning. However, despite its growing importance, structured and systematic surveys of multilingual prompt engineering methods remain limited. In this work, we provide an extensive survey of multilingual prompting strategies, categorizing them based on the NLP tasks they address. We examine 36 research papers published in the last two years, covering 39 prompting techniques applied to 30 multilingual NLP tasks. Our study presents a taxonomy diagram, tabulates the prompting techniques tested on different multilingual datasets cumulatively covering around 250 languages, analyzes the LLMs employed and identifies the potential SoTA method for each dataset. Furthermore, we conduct a comprehensive analysis across language families and resource classifications (high-resource vs. low-resource), exploring trends such as how NLP tasks are distributed based on resource availability and how prompting strategies vary across language families.

Existing surveys \cite{vatsal2024survey, sahoo2024systematic, edemacu2024privacy, chen2023unleashing, liu2023pre, singh2024promptengineering} primarily focus on generic prompt engineering methods rather than specifically addressing multilingual settings. While \citet{schulhoff2024prompt} briefly touches upon multilingual prompt engineering, it does not explore this dimension in the depth or breadth undertaken in our work, particularly with regard to its relationship with NLP tasks, language families and language resource levels. Similarly, \citet{huang2024survey, qin2024multilingual} review recent advances in multilingual LLMs rather than focusing on prompt engineering and its application to different NLP tasks. By structuring multilingual prompt engineering techniques according to NLP tasks, this survey aims to serve as a valuable resource for researchers and practitioners looking to develop more effective multilingual LLMs and corresponding prompt engineering techniques.

\subsection{Literature Search Process}

\footnotesize
\begin{longtable}{p{.6in}p{3.6in}}
    \caption{Google Scholar Queries for Selecting Papers in Literature Search Process}
    \label{tab:sq} \\
    \toprule
    \textbf{No.}    & \textbf{Query}  \\
    \midrule
    \endfirsthead
    
    \multicolumn{2}{c}%
    {{\bfseries \tablename\ \thetable{} -- continued from previous page}} \\
    \toprule
    \textbf{No.}    & \textbf{Query}\\
    \midrule
    \endhead

    \midrule \multicolumn{2}{r}{{Continued on next page}} \\
    \endfoot
    
    \bottomrule
    \endlastfoot

        1 & Prompt Engineering for Multilingual NLP Tasks  \\
        \midrule
        2    & Is Prompt Engineering Same for Different Languages  \\
        \midrule
        3     & Prompt Engineering for Different Languages\\
        \midrule
        4 & Prompt Engineering for Cross-lingual NLP Tasks  \\
        \midrule
        5    & Multilingual Prompt Engineering for Low-Resource Languages  \\
        \midrule
        6     & Prompt Engineering for Machine Translation  \\
        \midrule
        7 & Cross-lingual Prompt Engineering for Low-Resource Languages  \\
        \midrule
        8    & A Survey of Multilingual Prompt Engineering \\
        \midrule
        9    & A Survey of Prompt Engineering Methods for Low-Resource Languages \\
        \midrule
        10 & Prompt Engineering for Code-Switching  \\
        \midrule
        11    & Prompt Engineering in Multilingual Setting  \\

\end{longtable}

We use a two-step filtering process to finalize the set of research papers reviewed in our work.

Step 1: We perform basic search queries on Google Scholar\footnote{https://scholar.google.com} to compile an initial collection of research articles focused on prompt engineering in multilingual settings. In total, we use 11 search queries, listed in Table \ref{tab:sq}. After applying manual filtering, this results in a collection of 189 research articles.

Step 2: We apply two selection criteria to arrive at a final set of 36 research papers for our review. The first criterion requires that the selected articles explore prompt engineering primarily with hard prompts rather than soft prompts. A hard prompt consists of tokens corresponding to words in the LLM vocabulary, whereas a soft prompt involves parameter tuning of the LLM, using tokens that may not correspond to the LLM vocabulary but exist in a continuous space \cite{lester2021power,wang2023review}. The second criterion specifies that an article is included in this survey only if the prompt engineering method discussed has been applied to at least two languages.

\subsection{Outline}

The rest of the paper is organized in the following way. Section 2 talks about various prompt engineering techniques and section 3 highlights different NLP tasks. The sub-sections of section 3 discuss different prompting strategies that have been applied on a given NLP task and their corresponding results. Section 4 presents key insights into NLP tasks and prompting techniques, examining their associations with factors such as language resource availability and language family distribution. Finally, section 5 concludes the paper.

\section{Prompt Engineering Techniques}

In this section, we provide a brief overview of various prompting methods and how they have contributed to performance improvements over time. Notably, many of these prompting strategies have been explored in different settings or variations while maintaining the same core concept. The most common of these settings include zero-shot and few-shot. In the zero-shot setting \cite{radford2019language}, an LLM performs a task solely based on prompt instructions, without any training data, relying entirely on the knowledge it acquired during pre-training. Conversely, in the few-shot or in-context setting \cite{brown2020language}, the model is given a small number of task-specific examples alongside the prompt to aid comprehension. While research has shown that the few-shot approach often leads to better performance, it comes with the challenge of curating examples, as LLMs can exhibit unexplained biases toward the selected few-shot datapoints.

As we examine various multilingual prompt engineering methods, we adopt several key standardization practices. When variants of a core prompting strategy do not represent a substantial deviation from the original approach, we group them under a single overarching technique rather than treating them as distinct methods. Specifically, we consolidate such variants under the notation \{Prompting Technique\} + Variations \ref{variation}, where the \{Prompting Technique\} placeholder is replaced by the actual name of the primary prompting technique.   To ensure consistency across works that employ similar strategies, we have also standardized the terminology by renaming certain prompting methods—often differing from the original names used in the respective source papers. For instance, we refer to the method in \cite{shi2022language} as Translate-En-CoT to distinguish it from Translate-En in \cite{huang2023not}. Likewise, we have introduced the names Regressive En-CoT \ref{reg-eng-cot} and Regressive Native-CoT \ref{reg-native-cot} for strategies that correspond to CoT-based prompting techniques described in \cite{dwivedi2024navigating}. Similarly, all variants of Basic/Standard/Vanilla/Direct prompting across different papers have been unified under the terms En-Basic \ref{en-basic} or Native-Basic \ref{native-basic}, depending on the language used for instructing the LLM to solve the given task.

\subsection{En-Basic/Standard/Vanilla/Direct Prompting}

\label{en-basic}

Basic prompting refers to the approach of directly presenting a query to an LLM without any enhancements or modifications intended to boost performance—improvements that are typically the focus of most advanced prompting strategies. A basic prompt may include simple elements such as role assignments, a detailed task description, specifications of the desired output format and other straightforward instructions to guide the LLM’s response. In the literature, this technique is often referred to interchangeably as Standard prompting, Vanilla prompting or Direct prompting. In multilingual or cross-lingual settings, En-Basic denotes the variant where the instructions given to the LLM are in English. Several studies have also explored the few-shot version of this technique, wherein the exemplars included are also presented in English. Numerous prior studies \cite{huang2023not, lu2023chain, ghazvininejad2023dictionary, puduppully2023decomt, pilault2023interactive} have employed this strategy across a wide range of NLP tasks, as it often serves as the baseline against which newer prompting techniques are evaluated.

\subsection{Native-Basic/Standard/Vanilla/Direct Prompting}

\label{native-basic}

Native-Basic follows the same principles as En-Basic, with the key distinction that the instructions provided to the LLM are in a language other than English. Whenever the task instructions are presented in a language other than English, we classify that instance of Basic prompting as Native-Basic. Similar to En-Basic, the few-shot version of Native-Basic includes exemplar datapoints in the corresponding native language. Several studies have established their baselines using Native-Basic prompting methods, including \cite{shi2022language, qin2023cross, intrator2024breaking, trautmann2022legal}. Performance comparisons between Native-Basic and En-Basic vary depending on the specific NLP task. For example, \cite{asai2023buffet} reports that Native-Basic outperforms En-Basic in tasks such as Emotion/Sentiment Understanding and Coreference Resolution. Conversely, \cite{liu2024translation} finds that En-Basic performs better for tasks like Mathematical Problem Solving, Causal Reasoning, Natural Language Inference, Paraphrasing, Context-Free Question Answering and Summarization.

\subsection{X-Basic/Standard/Vanilla/Direct Prompting}

\label{x-basic}

X-Basic is inherently a few-shot version of Basic prompting which incorporates a factor of cross-linguality in a way where the few-shot datapoints and task solving instructions are in two different languages with one of them being in English and the other one being a native language.

\subsection{Native-CoT}

\label{native-cot}

\citet{shi2022language} proposes this prompting method where in a multilingual setting, CoT is employed to solve the problem in the native language by generating the reasoning steps in the original language of the problem. The authors suggest that this approach enables an evaluation of the model’s ability to both comprehend and solve tasks in a given language. Native-CoT is seen to achieve on average, approximately 30\% higher performance than Basic prompting across both high-resource as well as low-resource languages for Mathematical Problem Solving task. 

\subsection{En-CoT}
In contrast to Native-CoT, En-CoT \cite{shi2022language} generates reasoning steps in English, regardless of the original or native language of the problem. The rationale behind En-CoT is that English is frequently used as a source language in cross-lingual transfer and has been shown to be effective when employed as the prompt language. En-CoT outperforms Basic prompting and Native-CoT by an average of 33\% and 3\%, respectively, in the Mathematical Problem Solving task. For the Causal Reasoning task, En-CoT achieves approximately 6\% higher performance than Basic prompting.

\subsection{Translate-En-CoT}

Building upon the concepts of Native-CoT and En-CoT, the Translate-En-CoT prompting method \cite{shi2022language} involves translating the problem from its original language into English prior to generating the reasoning steps. This approach is based on the premise that, since English is the predominant language in LLM training, translating the problem into English can enhance the model's performance in solving it. The results show that Translate-En-CoT performs better than Basic, Native-CoT and En-CoT by approximately 37\%, 11\% and 4\% respectively, in Mathematical Problem Solving task. Translate-En-CoT can be implemented in two distinct settings. In one setting, the same LLM used for solving the task is also responsible for performing the translation. In the alternative setting, an external Machine Translation system is employed for the initial translation step.

\subsection{Translate-En}

The Translate-En prompting technique \cite{huang2023not} comprises two sequential steps. The first involves translating the problem from its original language into English, while the second step entails solving the translated problem in English. Translate-En consistently outperforms or yields comparable results to Basic prompting across a range of tasks, including Causal Reasoning, Natural Language Inference, Paraphrasing and Mathematical Problem Solving. Similar to Translate-En-CoT, Translate-En can be implemented in two distinct configurations. In one setting, the same LLM used for solving the task is also employed to perform the translation. In the alternative setting, an external Machine Translation system is utilized for the initial translation step.


\subsection{Cross-Lingual-Thought Prompting (XLT)} 
The prompting strategy discussed in \citet{huang2023not} consists of five key components. First, the LLM is instructed to assume a specific role and is informed about the task language. Next, to encourage a cross-lingual thought process, the LLM translates the task input from the native language to English. Following translation, the LLM proceeds to solve the task. In the fourth step, the LLM generates a reasoning chain for the solved task. Finally, the LLM is instructed to follow a specified format when generating the output. XLT consistently outperforms En-CoT, Basic and Translate-En prompting methods across multiple datasets in Mathematical Problem Solving, Causal Reasoning, Natural Language Inference, Paraphrasing, Context-Free Question Answering, Summarization and Machine Translation, using various LLMs.


 \subsection{Cross-Lingual-Prompting (CLP)}
 
In this work \cite{qin2023cross}, the authors try to bridge the gap between languages for solving a given task using cross-lingual CoT. The CLP prompting technique consists of two primary steps. The first step focuses on cross-lingual alignment, where the LLM is instructed to comprehend the task and generate the corresponding reasoning chain in English rather than the native language. The authors suggest that this step enables the LLM to establish representational connections between the native language and English for the given task. In the second step, the LLM is directed to solve the task in English and generate the corresponding reasoning chain. Experimental results demonstrate that CLP outperforms Basic, Native-CoT, En-CoT and Translate-En across Mathematical Problem Solving and Causal Reasoning datasets in all evaluated languages. Additionally, CLP surpasses En-CoT in performance on Natural Language Inference and Paraphrasing datasets.


\subsection{Cross-Lingual Self-Consistent Prompting (CLSP)}

CLSP \cite{qin2023cross} builds upon CLP by introducing greater linguistic flexibility. In the first step of CLP, where the LLM is asked to understand the task in English, CLSP instead instructs the LLM to comprehend the task in a given language \textit{l}, which may differ from English and to employ the corresponding reasoning steps in that language. The second step remains unchanged from CLP, but rather than solving the task in a fixed language (English), the LLM is now prompted to generate the solution in the given language \textit{l}. Finally, the most consistent answer across these different language-based reasoning chains is selected. The empirical results show that CLSP consistently outperforms all its counterparts, including CLP, Basic, Native-CoT, En-CoT and Translate-En, across all evaluated languages except Telugu in the Mathematical Problem Solving task. Similarly, for the Causal Reasoning task, CLSP achieves superior results compared to CLP, Basic, Translate-En and En-CoT across all languages except Tamil.


\subsection{Semantic Alignment Prompting}
\label{Sem-Align}

\citet{tanwar2023multilingual} proposes Semantic Alignment prompting, a few-shot prompting technique that involves selecting semantically similar datapoints in the source language while instructing the LLM to solve a task in the target language. The authors hypothesize that despite linguistic differences between the few-shot examples and the test datapoint, the shared embedding space would still capture their underlying similarities. Additionally, the authors introduce a variation of Semantic Alignment prompting, called Random prompting, where the few-shot examples in the source language are selected randomly rather than based on semantic similarity. Experimental results demonstrate that Semantic Alignment prompting consistently outperforms Random prompting across all evaluated language pairs for Sentiment Understanding and Toxicity Understanding tasks.


\subsection{Task-Based Alignment Prompting}
Task-Based Alignment prompting \cite{tanwar2023multilingual} builds upon Random prompting by incorporating a manually designed statement, referred to as the task-aligner, which is appended to each randomly selected few-shot example in the source language. The task-aligner serves to guide the LLM in mapping the label space from the source language to the target language. Empirical results indicate that Task-Based Alignment prompting consistently outperforms Random prompting for both Sentiment Understanding and Toxicity Understanding tasks across nearly all evaluated language pairs. 


\subsection{Cross-Lingual In-Context Source-Target Alignment (X-InSTA)}

X-InSTA \cite{tanwar2023multilingual} combines the Semantic Alignment and Task-Based Alignment prompting methods. First, similar to the Semantic Alignment approach, semantically similar few-shot examples are selected in the source language. Then, following the Task-Based Alignment strategy, task-aligners are appended to these selected few-shot datapoints to facilitate label space mapping between languages. X-InSTA outperforms Task-Based Alignment, Random and Semantic Alignment prompting techniques, making it the best-performing method for both Sentiment Understanding and Toxicity Understanding tasks.

\subsection{Chain-of-Dictionary (CoD)}

The authors of \cite{lu2023chain} come up with a prompting strategy to address the challenge of selecting relevant examples for few-shot Machine Translation task, particularly for low-resource languages. CoD prompting technique integrates multilingual dictionary information as prior knowledge directly into the conventional translation prompt. When given a test datapoint, a subset of words is selected and their corresponding multilingual dictionary entries are retrieved. Before making the standard translation request to the LLM, additional textual inputs outlining possible chained multilingual translations for these words are appended to the prompt. This method provides the model with explicit cross-linguistic mappings, aiding in more accurate translations. Statistical results presented by the authors show that CoD improves Machine Translation performance compared to Basic prompting in 83.75\% of translation directions across 200 languages. The authors further experimented with variations of CoD, replacing chained dictionaries with monolingual, multilingual and decomposed dictionaries; however, none of these alternatives surpassed the performance of CoD.


\subsection{Dictionary-based Prompting for Machine Translation (DIPMT)}

\label{dipmt}

\citet{ghazvininejad2023dictionary} comes up with another dictionary-based prompting technique for Machine Translation, specifically addressing the challenges associated with low-resource language and out-of-domain translations. Their approach, DIPMT, enhances translation quality by incorporating dictionary-based word mappings directly into the prompt. Given a test datapoint, dictionary entries for the words present in the sentence are retrieved and appended to the prompt. After requesting the translation of the full sentence, an additional section is included in the prompt, listing possible translations for specific words. This method provides the model with explicit lexical guidance, helping it generate more accurate translations. Empirical results indicate that DIPMT improves translation performance compared to Basic prompting, yielding an average increase of 1 BLEU point across both translation directions for 10 low-resource languages and an average improvement of 9.4 BLEU points across four out-of-domain experiments.


\subsection{Sequential Autoregressive Prompting (SAP)}

\label{sap}

Introduced in \citet{patel2022bidirectional}, this prompting method is based on the core idea of enabling few-shot learning in bidirectional language models. The authors conduct their experiments using mT5-3.7B (mT5-XL) \cite{xue2020mt5}. This technique involves iteratively prompting mT5 T times until the model generates a stop token. At each iteration, the first word generated by the model is retained (using a space character as a delimiter) and concatenated to the last line of the prompt for use in the next iteration. This approach enables mT5 to achieve performance levels comparable to unidirectional models such as XGLM and GPT-3, while utilizing approximately 50\% fewer parameters and requiring 16× fewer examples across 182 language pairs for the Machine Translation task. Additionally, for Contextual Question-Answering, SAP significantly outperforms the Basic prompting method, highlighting its effectiveness beyond translation tasks.

The authors also introduce a bootstrapped version of SAP, in which synthetic few-shot examples are generated in a completely unsupervised manner using a zero-shot translation prompt with no prior examples. The performance of this bootstrapped version is further improved through an ensembling strategy, where the top N synthetic examples are selected to form N/2 few-shot prompts, and the best translations are chosen using the mT5Score metric. Experimental results demonstrate that bootstrapped SAP outperforms the standard SAP by an average of 1\% across 186 language pairs in Machine Translation. 


\subsection{Decomposed Prompting for Machine Translation (DecoMT)}

The authors of \citet{puduppully2023decomt} draw inspiration from the notion that, for related languages, the task of Machine Translation can be simplified by utilizing the monotonic alignment characteristic inherent to such languages. The DecoMT method employs a two-stage translation process for word chunks: first, an independent translation stage, where each chunk is translated in isolation; and second, a contextual translation stage, where translation is performed while considering the surrounding context. The experimental results demonstrate that DecoMT outperforms both Basic prompting and SAP in translation quality across seven closely related language pairs, highlighting its effectiveness in leveraging language similarities for improved translation performance.


\subsection{Ambiguity Context}

The work presented in \cite{pilault2023interactive} introduces the Ambiguity Context prompting strategy, aimed at improving Machine Translation NLP task by explicitly incorporating contextual information alongside the sentence to be translated. This approach provides the LLMs not only with the target sentence but also with preceding discourse or surrounding sentences to assist in resolving ambiguities. The inclusion of context enables the model to better disambiguate meanings, preserve discourse continuity and accurately interpret linguistic subtleties such as formality, coreference and polysemy. Empirical results indicate that Ambiguity Context outperforms Basic prompting methods that lack additional contextual cues, thereby underscoring the critical role of surrounding text in achieving high-quality multilingual translation.

\subsection{Interactive-Chain-Prompting (InterCPT)}

The authors of \cite{pilault2023interactive} come up with InterCPT prompting technique where they try to solve a Machine Translation task by breaking it down into subproblems. This method consists of three key steps: (1) Identifying ambiguities, where potential issues in translation are recognized; (2) Resolving ambiguities, where the LLM addresses the subproblems identified in the first step; and (3) Combining the inputs and outputs from the first two steps to generate the final translation. The experimental results show that InterCPT is able to obtain better results than Basic prompting, a variation of Basic prompting where additional context is provided on top of Basic promoting, as well as Google Translate, an external Machine Translation system. These findings highlight the effectiveness of the InterCPT approach in improving translation quality by systematically addressing ambiguities during the translation process.


\subsection{Human-in-the-Loop (HIL)}

The HIL prompting strategy \cite{yang2023human} is based on the intuition that few-shot or in-context learning closely mirrors human cognitive abilities in NLP tasks. Building on this idea, the authors introduce HIL, a two-stage approach for Machine Translation, where the LLM refines its output iteratively. In the first stage, the LLM generates an initial draft translation, which serves as a preliminary version to be refined in the next step if necessary. Additionally, this stage involves the creation of a data store that collects human or automated feedback on historical translation errors in parallel. The second stage focuses on retrieving the most relevant datapoints from this data store, based on a predefined similarity metric, given a test datapoint for translation. The LLM then refines the draft translation from the first stage using the retrieved exemplars to improve performance. A variation of HIL introduces an additional third stage, where the LLM is prompted to compare and refine the translations produced in the first two stages, further enhancing translation quality. Experimental results indicate that HIL performs either comparably to or better than the Basic prompting strategy across five different domains in one evaluated dataset for Machine Translation.


\subsection{Rerank}

The prompting technique discussed in \citet{he2024exploring} closely resembles the Basic prompting method, utilizing the same prompt structure. However, it differs by generating multiple outputs through sampling at a lower temperature, allowing for variations in the responses. The results show that Rerank performs better than the Basic prompting strategy almost always in the Machine Translation task across 11 language pairs, highlighting the effectiveness of leveraging multiple generated outputs for improved translation quality.

\subsection{Multi-Aspect Prompting and Selection (MAPS)}

In \citet{he2024exploring}, the authors investigate whether LLMs can be guided to mimic professional human translators by taking preparatory steps before translating a given test datapoint. These steps involve gathering and analyzing relevant information, such as keywords, topics and example data, to improve translation quality. The proposed approach, MAPS, consists of three key steps. The first step of knowledge mining requires the LLM to analyze the test datapoint and extract three essential aspects of knowledge beneficial for translation: keywords that capture the core meaning, the overall topic, and equivalent few-shot demonstrations to provide contextual guidance. The second step of knowledge integration requires the LLM to incorporate the extracted knowledge from step one and generates multiple translation candidates while considering the given context. The last step of knowledge selection mirrors the decision-making process in human translation, where the model selects the best translation of the test datapoint's text based on the contextual information. MAPS consistently enhances the translation capabilities of LLMs, outperforming both Rerank and Basic prompting strategies across 11 language pairs. These findings highlight the effectiveness of a structured, knowledge-driven approach in improving LLM-based translation.

\subsection{ChatIE} 

\citet{wei2023chatie} focuses on a technique that reformulates the zero-shot information extraction task as a multi-turn question-answering problem using a two-stage framework. In the first stage, the LLM identifies the types of entities, relations or events present in the given test datapoint through a single question-answering turn. In the second stage, the model extracts the corresponding entities for the identified entity types, typically requiring multiple question-answering turns. Empirical results demonstrate that ChatIE significantly outperforms the Basic prompting strategy, achieving up to a 60\% improvement in Relation Extraction task, up to 5\% in Named Entity Recognition task and up to 3\% in Event Extraction task.


\subsection{Translator + Post Edit Prompting}

\citet{raunak2023leveraging} introduces Translator + Post Edit prompting, a method for Machine Translation that consists of two key steps. In the first step, an external translation system generates an initial translation for a given test datapoint. In the second step, the LLM refines and improves upon the initial translation. The authors demonstrate that this translation post-editing approach, applied in a direct setting without any quality estimation or error detection step prior to post-editing, always outperforms the use of an external translator or Basic prompting alone. Empirical results show that Translator + Basic prompting achieves superior translation quality across three datasets and four language pairs.

\subsection{Translator + Post Edit CoT Prompting}

Building upon Translator + Post Edit prompting, Translator + Post Edit CoT prompting \cite{raunak2023leveraging} introduces an additional step where the LLM first generates potential edits or improvements before applying these refinements to produce an enhanced translation. The authors also present a variation of this technique, referred to as Translator + Post Edit Structured-CoT (SCoT), which differs in its use of the Multidimensional Quality Metrics (MQM) Framework \cite{freitag2021experts}. This approach incorporates MQM annotation instructions to generate the intermediate CoT as an MQM-style annotation over the source-translation pair. The experiments conducted by authors show that both Translator + CoT and Translator + Structured-CoT achieve performance levels comparable to the Basic prompting method while outperforming the use of an external translator alone across three evaluated datasets.


\subsection{Fuzzy Match}

\citet{moslem2023adaptive} discusses a few-shot Machine Translation prompting approach in which the few-shot examples are dynamically selected from a translation memory based on embedding similarity. This ensures that the selected examples are highly relevant to the test datapoint, as they are chosen based on linguistic similarity rather than being manually curated or randomly selected. Conceptually, this technique is closely related to Semantic Alignment prompting \ref{Sem-Align}; however, while Semantic Alignment is applied to a different NLP task, Fuzzy Match is specifically designed for Machine Translation. Experimental results demonstrate that Fuzzy Match consistently outperforms the Basic prompting technique across all evaluated language pairs.

\subsection{Translator + Fuzzy Match}

Building upon the Fuzzy Match approach, \citet{moslem2023adaptive} introduces Translator + Fuzzy Match, which integrates external Machine Translation into the prompting process. The authors explore two experimental settings. In the first setting, in addition to retrieving fuzzy matches, an external Machine Translation of the test datapoint is appended before prompting the LLM to generate the final translation. In the second setting, translations of all fuzzy matches are appended alongside the retrieved fuzzy matches before prompting the LLM to generate the final translation. Empirical results demonstrate that Translator + Fuzzy Match outperforms both the external translator alone and the Fuzzy Match strategy alone across all evaluated language pairs. Furthermore, among the two settings, the second setting, where translations of all fuzzy matches are incorporated, achieves better results than the first setting, which only includes the translation of the test datapoint.

\subsection{Terminology + Fuzzy Match}

Another variation of the Fuzzy Match prompting strategy, introduced by \citet{moslem2023adaptive}, is Terminology + Fuzzy Match, which enhances the translation process by incorporating terminological resources. The authors investigate two settings. In the first setting, in addition to fuzzy matches, the prompt includes n-gram terms extracted from the fuzzy matches in both the source and target languages. The LLM is then prompted to generate the final translation based on this enriched context. In the second setting, in addition to the n-gram terms from fuzzy matches, additional terms from a glossary compiled by human annotators are incorporated to further enhance the translation context. Experimental results indicate that Terminology + Fuzzy Match always does better than the Fuzzy Match technique across all evaluated language pairs, demonstrating the effectiveness of integrating domain-specific terminology into the prompting process.

\subsection{Multi-Turn Dialogue}

The authors of \citet{wu2023exploring} talk about Multi-Turn Dialogue prompting technique for Machine Translation that involves a step-by-step translation process. Unlike Basic prompting, where the LLM translates the entire test datapoint in a single pass, this method instructs the model to translate sentence by sentence in an interactive manner. Experimental results show that Multi-Turn Dialogue prompting achieves superior performance compared to Basic prompting, which requires the LLM to translate the entire input in one step. This performance improvement is consistently observed across both evaluated language pairs, English-Chinese and Chinese-English.


\subsection{Iterative Prompting}

\citet{nie2024decomposed} investigates Iterative Prompting in a multilingual setting for sequence labeling tasks. In this approach, the LLM is prompted iteratively, predicting the entire test datapoint step by step. At each timestep, the predicted label is appended to the context along with the next word, allowing the model to generate labels sequentially. The results presented in \citet{nie2024decomposed} indicate that Iterative Prompting yields inferior performance compared to Decomposed Prompting.

\subsection{Decomposed Prompting (Generative/Probabilistic)}

\label{decomposed}
Decomposed Prompting \cite{nie2024decomposed} is a prompting strategy for sequence labeling NLP tasks, inspired by the step-by-step reasoning process humans employ when annotating linguistic features in a sentence. This method mirrors human annotation by first decomposing a given test datapoint into individual tokens. A separate prompt is then generated for each token, effectively transforming the sequence labeling task into a series of focused, token-specific prompts. This approach is evaluated under two settings. Probability-based setting – this method leverages the model’s output logits to compute the probability distribution over the sequence labeling tag set, assigning the label with the highest probability. Generation-based setting – this method directly compares the LLM-generated content with the gold label to determine the correct classification. Experimental results indicate that the probability-based approach consistently outperforms the generation-based setting in sequence labeling tasks. Furthermore, empirical findings confirm that Decomposed Prompting significantly outperforms Iterative Prompting, with the largest observed performance gap reaching up to 17\%.


\subsection{Translate After LEarNing Textbook (TALENT)}

The Translate After LEarNing Textbook (TALENT) \cite{guo2024teaching} method is a prompting strategy designed to enhance LLMs' translation capabilities for low-resource languages. Inspired by human language acquisition, TALENT guides LLMs through a structured learning process akin to studying a textbook. The approach involves three key steps: (1) constructing a textbook that introduces the low-resource language's syntax patterns; (2) prompting the LLM to absorb and internalize these patterns; and (3) leveraging the acquired knowledge to improve translation performance. Empirical evaluations across 112 low-resource languages demonstrate that TALENT consistently improves translation quality, achieving a 14.8\% enhancement over zero-shot baselines. These findings suggest that structured, textbook-based prompting can effectively augment LLMs' understanding and generation capabilities in low-resource language contexts.

\subsection{Regressive En-CoT}

\label{reg-eng-cot}

\citet{dwivedi2024navigating} proposes Regressive En-CoT, a prompting approach designed for Subjectivity Analysis NLP task. In this method, the LLM is first instructed to assign a subjectivity score to a given test datapoint, indicating the degree of subjectivity present. Within the same prompt, the model is further required to justify the assigned score by providing a reasoning explanation. Since the entire prompt is formulated in English, the approach is referred to as Regressive En-CoT. Empirical results indicate that Regressive En-CoT performs worse than Regressive Native-CoT across all nine evaluated languages.


\subsection{Regressive Native-CoT}

\label{reg-native-cot}

Similar to Regressive En-CoT \cite{dwivedi2024navigating}, this method employs the same two-step prompting approach but replaces English with the native language of the test datapoint. The results reported by the authors demonstrate that Regressive Native-CoT outperforms Regressive En-CoT across all nine evaluated languages in the Subjectivity Analysis task. Among the evaluated languages, Telugu exhibits the most significant improvement across the chosen evaluation metrics. In contrast, Bengali, Maithili, and Urdu show the least performance gains.

\subsection{Human Directed (HD) Prompting}

\label{hd}

\citet{ferron2023meep} introduces the HD prompting approach, based on the intuition that LLMs, designed to approximate natural language understanding and human-like linguistic interactions, should benefit from prompts that mirror the instructions provided to human annotators. The authors hypothesize that structuring evaluation prompts in a manner similar to those used in human annotation will enhance model performance compared to Basic prompting. Empirical results demonstrate that HD prompting consistently outperforms Basic prompting in the Dialogue Evaluation task across all three evaluated LLMs.

\subsection{Metric for Engagingness
Evaluation using Prompting (MEEP)}

\label{meep}

Building on top of HD prompting technique, the authors of \citet{ferron2023meep}  propose MEEP where they add short phrases for their defined five sub-dimensions of engagingness including response diversity, interactional quality, interestingness, contextual specificity and othering. Additionally, the authors present an alternative version of MEEP, where the default zero-shot approach is transformed into a few-shot setting by incorporating example dialogue responses using the keyword "such as." Experimental results indicate that the few-shot version consistently outperforms the zero-shot version. Furthermore, MEEP achieves superior performance compared to Basic, HD, G-Eval and GPTScore prompting techniques in Dialogue Evaluation across all three evaluated languages: English, Spanish and Chinese.

\subsection{G-Eval}

\label{g-eval}

Originally introduced by \citet{liu2023g} and later evaluated in a multilingual setting by \citet{ferron2023meep}, G-Eval is a prompting strategy that comprises three key components: (1) a prompt that defines the evaluation task and specifies the desired evaluation criteria, (2) a CoT reasoning process, where the LLM generates a sequence of intermediate instructions outlining the detailed evaluation steps (3) a scoring function, which calls the LLM and computes a score based on the probabilities of the generated tokens. The reported results demonstrate that, in multilingual settings, G-Eval consistently outperforms GPTScore across all experiments. However, its overall performance remains inferior to MEEP, indicating that further refinements may be necessary to enhance its effectiveness in Dialogue Evaluation tasks.

\subsection{GPTScore}

\label{gptscore}

First proposed by \citet{fu2023gptscore} and later evaluated in a multilingual setting by \citet{ferron2023meep}, GPTScore is based on the principle that a generative pre-trained model assigns higher probabilities to high-quality generated text when following a given instruction and context. The instruction in GPTScore consists of two key components: the NLP task description and the aspect definition, where the aspect definition specifies the specific evaluation criteria for assessing the generated text. GPTScore is formulated as a weighted conditional probability, integrating these elements to quantify text quality. The results from \citet{ferron2023meep} indicate that GPTScore underperforms compared to both MEEP and G-Eval in Dialogue Evaluation tasks.

\subsection{ \{Prompting Technique\} + Variations}


When variations of a core prompting approach do not constitute a significant methodological departure from the original method, we consolidate them under a single overarching technique rather than treating them as independent methods. These variants are represented using the notation {Prompting Technique} + Variations, where {Prompting Technique} is replaced by the name of the underlying strategy. This practice allows for clearer comparisons across works and reduces fragmentation in the classification of prompting methodologies. For instance, in Table \ref{tab:cmemotion}, one of the prompting methods observed across nearly all datasets is En-Basic + Variations, where the core technique is En-Basic \ref{en-basic}. This enables the aggregation of related results and provides a more holistic view of how foundational approaches evolve across datasets. Similarly, in Table \ref{tab:mtrans}, for the FLORES-101 dataset, prompting methods such as DIPMT + Variations and SAP + Variations were evaluated, corresponding to the core strategies DIPMT \ref{dipmt} and SAP \ref{sap}, respectively. By standardizing the representation of method variants in this manner, we promote consistency and facilitate more meaningful analysis of methodological trends in multilingual prompt engineering research.

\label{variation}

\section{Prompt Engineering on Different Multilingual NLP Tasks}

\begin{figure}[!ht]
\begin{forest}
for tree={
    grow=east,
    draw,
    rounded corners,
    align=center,
    text width=4cm,
    inner xsep=4pt,
    inner ysep=2pt,
    l sep=5mm,
    s sep=1mm,
    parent anchor=east,
    child anchor=west,
    anchor=west,
    calign=first,
    edge path={
        \noexpand\path [draw, \forestoption{edge}] (!u.parent anchor) -- +(3mm,0) |- (.child anchor)\forestoption{edge label};
    },
    font=\sffamily\scriptsize,
    if level=0{
            fill=red!20,
            text width=1.1cm, 
            inner xsep=6pt, 
            inner ysep=4pt,  
            align=center,
            text centered
        }{},
    if level=1{
            text width=2.3cm, 
            inner xsep=5pt, 
            inner ysep=3pt,  
            align=center,
            text centered
        }{},
    if level=2{
            text width=2.8cm, 
            inner xsep=4pt, 
            inner ysep=2pt,  
            align=center,
            text centered
        }{},
    if level=3{
            text width=6cm, 
            inner xsep=4pt, 
            inner ysep=2pt,  
            align=center,
            text centered
        }{}
},
[NLP Tasks
    [Text Classification, fill=orange!20
        [Native-Basic, fill=orange!20
        [{
        German, French, Italian, English \\  \cite{trautmann2022legal}
        }, fill=orange!20]
        ]
    ]
    [Hope Detection, fill=green!20
        [{En-Basic, En-Basic + \\ Variations, En-CoT }, fill=green!20
        [{
        English, Spanish \cite{thuy2024empirical}
        }, fill=green!20]
        ]
    ]
    [Social Bias, fill=yellow!20
        [{Native-Basic, Translate-En }, fill=yellow!20
        [{
        German, French, Spanish, Arabic, Hebrew, Italian, \\ Ukrainian, Russian \cite{ahuja2023mega}
        }, fill=yellow!20]
        ]
    ]
    [Paraphrasing, fill=cyan!20
        [{Native-CoT, \\ Translate-En-CoT, \\ En-Basic, En-CoT, \\ Translate-En,  XLT, CLP, \\ Native-Basic, X-Basic} , fill=cyan!20
        [{
        Spanish, French, German, Chinese, Japanese, \\ English, Korean \cite{etxaniz2023multilingual}, \\ \cite{huang2023not},  \cite{qin2023cross}, \\ \cite{liu2024translation}, \cite{asai2023buffet}, \cite{ahuja2023mega}
        }, fill=cyan!20]
        ]
    ]
    [Code-Mixing\\Summarization, fill=white!20
        [{En-Basic, En-Basic + \\ Variations}, fill=white!20
        [{
        Hinglish-English \cite{zhang2023multilingual}
        }, fill=white!20]
        ]
    ]
    ]   
\end{forest}
\caption{Taxonomy Diagram of Prompt Engineering Methods and Languages Applied Across Different Multilingual NLP Tasks}
\label{taxonomy}
\end{figure}
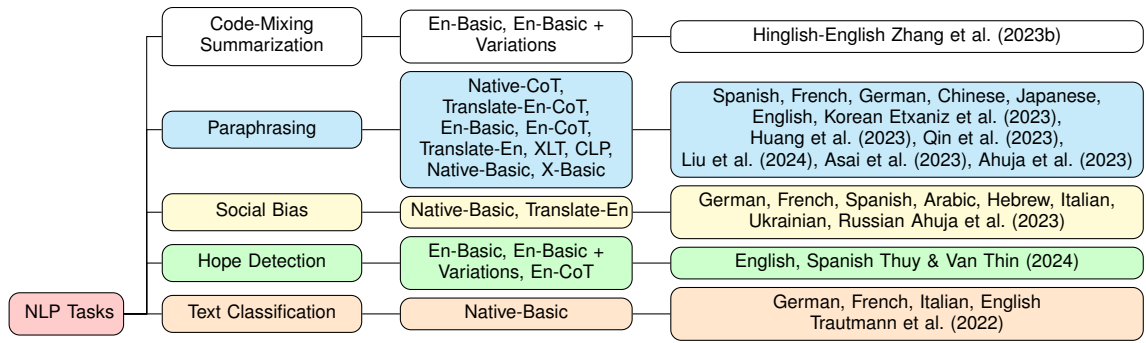

Various research studies have employed different criteria for categorizing datasets within multilingual NLP tasks, leading to inconsistencies across works. In this section, we aim to establish a standardized framework by defining distinct multilingual NLP tasks and systematically assigning datasets to these tasks. Additionally, we explore various prompting techniques that have been utilized for these tasks, along with the languages in which they have been tested. This unified approach not only facilitates direct comparisons between methodologies but also supports more transparent benchmarking of multilingual models. A taxonomy diagram illustrating this categorization is provided in Figure \ref{taxonomy}. It is important to acknowledge that certain datasets may be relevant to multiple NLP tasks simultaneously. However, assigning datasets to multiple categories can introduce complexities in analyzing the performance of prompting techniques across different multilingual tasks. Such overlap may also lead to ambiguous conclusions regarding the efficacy of specific methods. To maintain clarity, we ensure each dataset is associated with a single NLP task that best represents its primary function.

The following subsections define each multilingual NLP task, list the relevant datasets, describe the prompting methods applied, specify the number of unique languages or language pairs (for translation and code-mixing tasks) covered across all prompting methods and cite the corresponding research articles. This count reflects the union of languages or pairs where at least one prompting method has been applied. While individual methods may only cover subsets, the reported count captures the collective multilingual reach. We also identify possible SoTA prompting methods per dataset. Since effectiveness can vary by LLM, we provide a list of models used. However, when designating the SoTA strategy, we mention only the prompting method, as not all strategies have been tested across all models, making LLM-specific conclusions uncertain. Evaluation metrics are omitted, as they differ across studies. Additionally, the use of varying versions of the same dataset further complicates direct performance comparisons. Given these limitations, we rely on informed judgment to identify the most effective prompting method per dataset based on existing literature and multilingual results.

\begin{forest}
for tree={
    grow=east,
    draw,
    rounded corners,
    align=center,
    text width=4cm,
    inner xsep=4pt,
    inner ysep=2pt,
    l sep=5mm,
    s sep=1mm,
    parent anchor=east,
    child anchor=west,
    anchor=west,
    calign=first,
    edge path={
        \noexpand\path [draw, \forestoption{edge}] (!u.parent anchor) -- +(3mm,0) |- (.child anchor)\forestoption{edge label};
    },
    font=\sffamily\scriptsize,
    if level=0{
            fill=red!20,
            text width=1.1cm, 
            inner xsep=6pt, 
            inner ysep=4pt,  
            align=center,
            text centered
        }{},
    if level=1{
            text width=2.3cm, 
            inner xsep=5pt, 
            inner ysep=3pt,  
            align=center,
            text centered
        }{},
    if level=2{
            text width=2.8cm, 
            inner xsep=4pt, 
            inner ysep=2pt,  
            align=center,
            text centered
        }{},
    if level=3{
            text width=6cm, 
            inner xsep=4pt, 
            inner ysep=2pt,  
            align=center,
            text centered
        }{}
}
[NLP Tasks
    [Code-Mixing \\ Emotion/Sentiment \\ Understanding, fill=orange!20
        [{En-Basic, En-Basic + \\Variations, Native-Basic}, fill=orange!20
        [{
        Spanish-English, Malayalam-English, Tamil-English, \\English-Spanish [\cite{zhang2023multilingual}, \\\cite{ahuja2023mega}]
        }, fill=orange!20]
        ]
    ]
    [Emotion/Sentiment \\ Understanding, fill=white!20
        [{Random Prompting,\\ Semantic Alignment,\\ Task-Based Alignment,\\ X-InSTA, X-Basic,\\ Native-Basic, En-Basic}, fill=white!20
        [{
        German, English, Spanish, French, Japanese, Mandarin,\\ Assamese, Bengali, Boro, Gujarati, Hindi, Kannada,\\ Maithili, Malayalam, Marathi, Oriya, Punjabi, Tamil,\\ Telugu, Urdu, Chinese [\cite{tanwar2023multilingual}, \\\cite{asai2023buffet}]
        }, fill=white!20]
        ]
    ]
    [Subjectivity Analysis, fill=cyan!20
        [{Regressive En-CoT, \\Regressive Native-CoT}, fill=cyan!20
        [{
        Bengali, Gujarati, Hindi, Kannada, Maithili, Marathi, \\Tamil, Telugu, Urdu [\cite{dwivedi2024navigating}]
        }, fill=cyan!20]
        ]
    ]
    [Context-Free \\ Question-Answering, fill=yellow!20
        [{En-Basic, En-CoT, XLT, \\Native-Basic, Native-CoT, \\Translate-En-CoT}, fill=yellow!20
        [{
        Assamese, Bengali, Chinese, English, French, German, \\Gujarati, Hindi, Japanese, Kannada, Malayalam, \\Marathi, Odia, Punjabi, Russian, Spanish, Tamil, \\Telugu, Thai, Turkish, Vietnamese \cite{huang2023not}, \\\cite{liu2024translation}, \cite{lai2023chatgpt}]
        }, fill=yellow!20]
        ]
    ]
    [Contextual Question- \\ Answering, fill=green!20
        [{En-Basic, Native-Basic,\\ X-Basic, Translate-En, \\SAP}, fill=green!20
        [{
        Afrikaans, Amharic, Arabic, Armenian, Assamese, \\Bambara, Basque, Belarusian, Bengali, \\ Bengali (Romanized), Burmese, Catalan,\\ Central Kurdish, Cebuano, Chinese, \\Chinese (Simplified), Chinese (Traditional), Croatian, \\Czech, Danish, Dutch, Egyptian, English, Estonian,\\ Finnish, French, Georgian, German, Greek,\\ Gujarati, Haitian Creole, Halh Mongolian, Hausa,\\ Hebrew, Hindi, Hindi (Romanized), Hungarian,\\ Icelandic, Igbo, Ilocano, Indonesian, Italian,\\ Japanese, Javanese, Kannada, Kazakh, Khmer,\\ Kinyarwanda, Korean, Kurdish, Kyrgyz, Lao, Lingala,\\ Lithuanian, Macedonian, Malagasy, Malayalam, \\Maltese, Maori, Marathi, Mesopotamian Arabic, \\Modern Standard Arabic, Modern Standard Arabic \\(Romanized), Najdi Arabic, Nepali, Nepali (Romanized)\\, North Azerbaijani, North Levantine Arabic,\\ Northern Sotho, Northern Uzbek, Norwegian Bokmål, \\Odia, Plateau Malagasy, Polish, Portuguese, Punjabi,\\ Romanian, Russian, Serbian, Shona, Sinhala, Sindhi,\\ Slovak, Slovenian, Somali, Southern Pashto,\\ Southern Sotho, Spanish, Standard Latvian,\\ Standard Malay, Sundanese, Swahili, Swedish, Tajik,\\ Tamil, Telugu, Thai, Tigrinya, Tsonga, Turkish,\\ Ukrainian, Urdu, Urdu (Romanized), Vietnamese,\\ West Central Oromo, Western Persian, Wolof,\\ Xhosa, Yoruba, Zulu [\cite{kim2023cross}, \\\cite{patel2022bidirectional}, \cite{lai2023chatgpt}, \cite{ahuja2023mega}, \\\cite{intrator2024breaking}, \cite{asai2023buffet}]
        }, fill=green!20]
        ]
    ]
    [Causal Reasoning, fill=violet!20
        [{Native-Basic,  En-CoT, \\Native-CoT, En-Basic, \\Translate-En-CoT, \\Translate-En, XLT, CLP, \\CLSP, X-Basic}, fill=violet!20
        [{
        Chinese, Cusco-Collao Quechua, Estonian, Haitian,\\ Haitian Creole, Indonesian, Italian, Mandarin Chinese,\\ Quechua, Southern Quechua, Swahili, Tamil, Thai,\\ Turkish, Vietnamese [\cite{shi2022language}, \\\cite{etxaniz2023multilingual}, \cite{huang2023not}, \\\cite{qin2023cross}, \cite{kim2023cross}, \cite{liu2024translation}, \\\cite{asai2023buffet}, \cite{intrator2024breaking}, \\\cite{ahuja2023mega}]
        }, fill=violet!20]
        ]
    ]
    [Commonsense \\ Reasoning, fill=blue!20
        [{Native-CoT, En-Basic,\\ Translate-En-CoT, \\Native-Basic, Translate-En, \\X-Basic}, fill=blue!20
        [{
        Arabic, Basque, Burmese, Chinese, Dutch, English, \\French, German, Hindi, Indonesian, Italian, Japanese, \\Polish, Portuguese, Russian, Spanish, Swahili, Telugu, \\Urdu, Vietnamese [\cite{etxaniz2023multilingual}, \\\cite{intrator2024breaking}, \cite{lai2023chatgpt}, \\\cite{ahuja2023mega}]
        }, fill=blue!20]
        ]
    ]
    [Mathematical \\ Problem Solving, fill=pink!20
        [{Native-Basic, Native-CoT,\\ En-CoT, Translate-En-CoT,\\ En-Basic, Translate-En,\\ XLT, CLP, CLSP}, fill=pink!20
        [{
        Bengali, Chinese, English, French, German, Japanese,\\ Russian, Spanish, Swahili, Telugu, Thai [\\\cite{shi2022language}, \cite{etxaniz2023multilingual}, \\\cite{huang2023not}, \cite{qin2023cross}, \cite{liu2024translation}]
        }, fill=pink!20]
        ]
    ]    
]
\end{forest}

\begin{forest}
for tree={
    grow=east,
    draw,
    rounded corners,
    align=center,
    text width=4cm,
    inner xsep=4pt,
    inner ysep=2pt,
    l sep=5mm,
    s sep=1mm,
    parent anchor=east,
    child anchor=west,
    anchor=west,
    calign=first,
    edge path={
        \noexpand\path [draw, \forestoption{edge}] (!u.parent anchor) -- +(3mm,0) |- (.child anchor)\forestoption{edge label};
    },
    font=\sffamily\scriptsize,
    if level=0{
            fill=red!20,
            text width=1.1cm, 
            inner xsep=6pt, 
            inner ysep=4pt,  
            align=center,
            text centered
        }{},
    if level=1{
            text width=2.3cm, 
            inner xsep=5pt, 
            inner ysep=3pt,  
            align=center,
            text centered
        }{},
    if level=2{
            text width=2.8cm, 
            inner xsep=4pt, 
            inner ysep=2pt,  
            align=center,
            text centered
        }{},
    if level=3{
            text width=6cm, 
            inner xsep=4pt, 
            inner ysep=2pt,  
            align=center,
            text centered
        }{}
}
[NLP Tasks
[Code-Mixing \\ Language \\ Identification, fill=orange!20
        [{En-Basic, En-Basic + \\Variations}, fill=orange!20
        [{
        English-Hindi, Modern Standard Arabic-Egyptian \\Arabic [\cite{zhang2023multilingual}]
        }, fill=orange!20]
        ]
    ]
[Code-Mixing \\ Machine Translation, fill=green!20
        [{En-Basic, En-Basic + \\Variations}, fill=green!20
        [{
        Hinglish-English [\cite{zhang2023multilingual}]
        }, fill=green!20]
        ]
    ]
[Machine Translation, fill=yellow!20
        [{Ambiguity Context, CoD, \\CoD + Variations, \\DecoMT, DIPMT, \\DIPMT + Variations, \\En-Basic, \\En-Basic + Variations,\\ En-CoT, Fuzzy Match,\\ Fuzzy Match + Glossary,\\ HIL, HIL + Variations,\\ InterCPT, MAPS,\\ Multi-Turn Dialogue,\\ Multi-Turn Dialogue +\\Variations,\\ Rerank, SAP,\\ SAP + Variations,\\ TALENT, \\Translator + Fuzzy Match, \\Translator + Post Edit,\\ Translator + Post Edit \\CoT, Translator + Post Edit \\SCoT, XLT}, fill=yellow!20
        [{English-FLORES\textsuperscript{*}, Czech-Ukrainian\textsuperscript{*}, Hindi-Marathi\textsuperscript{*},\\ French-Russian\textsuperscript{*}, Catalan-Hindi\textsuperscript{*}, Malayalam-Arabic\textsuperscript{*},\\ Korean-Tamil\textsuperscript{*}, Tamil-Finish\textsuperscript{*}, Hindi-Tamil\textsuperscript{*},\\ Tamil-Malayalam\textsuperscript{*}, German-Vietnamese,\\ Malayalam-Finish\textsuperscript{*}, Bulgarian-Swahili\textsuperscript{*},\\ Chinese-Bulgarian\textsuperscript{*}, Persian-Russian\textsuperscript{*},\\ Portuguese-Spanish\textsuperscript{*}, Russian-Arabic\textsuperscript{*},\\ Malayalam-Russian\textsuperscript{*}, Bulgarian-Finish\textsuperscript{*}, Finish-French\textsuperscript{*},\\ Tamil-Catalan\textsuperscript{*}, French-Chinese\textsuperscript{*}, Chinese-Malayalam\textsuperscript{*},\\ Chinese-Finish\textsuperscript{*}, Catalan-Finish\textsuperscript{*}, Hindi-Telugu\textsuperscript{*},\\ Swahili-German\textsuperscript{*}, Catalan-French\textsuperscript{*}, Arabic-Korean\textsuperscript{*},\\ German-French\textsuperscript{*}, Russian-Ukrainian\textsuperscript{*}, Chinese-English\textsuperscript{*},\\ Javanese-Thai, Russian-Hindi\textsuperscript{*}, Finish-German\textsuperscript{*},\\ Catalan-Malayalam\textsuperscript{*}, Chinese-German\textsuperscript{*},\\ German-Catalan\textsuperscript{*}, Arabic-Swahili\textsuperscript{*}, Korean-Catalan\textsuperscript{*},\\ Chinese-Arabic\textsuperscript{*}, Finish-Arabic\textsuperscript{*}, Russian-Swahili\textsuperscript{*},\\ Chinese-Swahili\textsuperscript{*}, Korean-Russian\textsuperscript{*}, Hindi-Malayalam\textsuperscript{*},\\ Chinese-Javanese, Hindi-Finish\textsuperscript{*}, Russian-German\textsuperscript{*},\\ Tamil-Bulgarian\textsuperscript{*}, Hindi-Arabic\textsuperscript{*}, Finish-Swahili\textsuperscript{*},\\ Tamil-Chinese\textsuperscript{*}, Bulgarian-Malayalam\textsuperscript{*}, Thai-Irish,\\ Tamil-Russian\textsuperscript{*}, English-Persian\textsuperscript{*}, Tamil-French\textsuperscript{*},\\ Arabic-English\textsuperscript{*}, English-Finish\textsuperscript{*}, Bulgarian-Hindi\textsuperscript{*},\\ German-Arabic\textsuperscript{*}, German-Hindi\textsuperscript{*}, Korean-Bulgarian\textsuperscript{*},\\ Gujrati-Hindi\textsuperscript{*}, French-Swahili\textsuperscript{*}, Tamil-German\textsuperscript{*},\\ Hindi-Swahili\textsuperscript{*}, Bulgarian-German\textsuperscript{*}, Bulgarian-Arabic\textsuperscript{*},\\ Swahili-Tamil\textsuperscript{*}, Indonesian-Malay\textsuperscript{*}, Chinese-Russian\textsuperscript{*},\\ Russian-Catalan\textsuperscript{*}, Malayalam-Korean\textsuperscript{*},\\ French-Malayalam\textsuperscript{*}, Chinese-Catalan\textsuperscript{*}, Arabic-French\textsuperscript{*},\\ Korean-Swahili\textsuperscript{*}, English-Filipino\textsuperscript{*}, Swahili-Malayalam\textsuperscript{*},\\ Chinese-Korean\textsuperscript{*}, Arabic-Catalan\textsuperscript{*},\\ Catalan-Swahili\textsuperscript{*}, French-Bulgarian\textsuperscript{*}, Hindi-French\textsuperscript{*},\\ French-Korean\textsuperscript{*}, Korean-Finish\textsuperscript{*}, German-Korean\textsuperscript{*},\\ Tamil-Arabic\textsuperscript{*}, Bulgarian-Russian\textsuperscript{*}, Norwegian-English\textsuperscript{*},\\ Hindi-Chinese\textsuperscript{*}, Catalan-Bulgarian\textsuperscript{*}, Russian-Finish\textsuperscript{*},\\ English-Malay\textsuperscript{*}, Korean-Hindi\textsuperscript{*}, German-Malayalam\textsuperscript{*},\\ Chinese-Thai [\cite{ghazvininejad2023dictionary},\\ \cite{guo2024teaching}, \cite{he2024exploring}, \cite{he2024prompting},\\ \cite{huang2023not}, \cite{iyer2023towards}, \cite{lu2023chain},\\ \cite{moslem2023adaptive}, \cite{patel2022bidirectional},\\ \cite{wu2023exploring},
        \cite{pilault2023interactive},\\ \cite{yang2023human}, \cite{pourkamali2024machine},\\ \cite{puduppully2023decomposed}, \cite{raunak2023leveraging}]
        }, fill=yellow!20]
        ]
    ]
[Code-Mixing \\ Generation, fill=purple!20
        [{En-Basic, En-Basic + \\Variations}, fill=purple!20
        [{
        English-Indonesian, English-Malay, English-Chinese, \\English-Tagalog, English-Vietnamese, English-Tamil, \\English-Singlish [\cite{yong2023prompting}]
        }, fill=purple!20]
        ]
    ]
    [Question Generation, fill=white!20
        [{X-Basic, Native-Basic, \\En-Basic}, fill=white!20
        [{
        Arabic, Bengali, Finnish, Indonesian, Swahili, Korean, \\Russian, Telugu [\cite{asai2023buffet}]
        }, fill=white!20]
        ]
    ]
    [Toxicity \\ Understanding, fill=cyan!20
        [{Random Prompting, \\Semantic Alignment, \\Task-Based Alignment,\\ X-InSTA, Native-Basic,\\ X-Basic, Translate-En}, fill=cyan!20
        [{
        English, Turkish, Portuguese, Russian, Spanish,\\ Italian, French [\cite{tanwar2023multilingual}, \cite{ahuja2023mega}]
        }, fill=cyan!20]
        ]
    ]
]
\end{forest}


 \let\thefootnote\relax\footnotetext{\textsuperscript{*} on language pairs for the Machine Translation task indicates that some or all of the prompting strategies were applied in both directions of translation. The 204 languages of FLORES dataset in English-FLORES\textsuperscript{*} pair can be found at \href{https://github.com/facebookresearch/flores/blob/main/flores200/README.md}{FLORES Dataset.} \url{}
 }

\begin{forest}
for tree={
    grow=east,
    draw,
    rounded corners,
    align=center,
    text width=4cm,
    inner xsep=4pt,
    inner ysep=2pt,
    l sep=5mm,
    s sep=1mm,
    parent anchor=east,
    child anchor=west,
    anchor=west,
    calign=first,
    edge path={
        \noexpand\path [draw, \forestoption{edge}] (!u.parent anchor) -- +(3mm,0) |- (.child anchor)\forestoption{edge label};
    },
    font=\sffamily\scriptsize,
    if level=0{
            fill=red!20,
            text width=1.1cm, 
            inner xsep=6pt, 
            inner ysep=4pt,  
            align=center,
            text centered
        }{},
    if level=1{
            text width=2.3cm, 
            inner xsep=5pt, 
            inner ysep=3pt,  
            align=center,
            text centered
        }{},
    if level=2{
            text width=2.8cm, 
            inner xsep=4pt, 
            inner ysep=2pt,  
            align=center,
            text centered
        }{},
    if level=3{
            text width=6cm, 
            inner xsep=4pt, 
            inner ysep=2pt,  
            align=center,
            text centered
        }{}
}
[NLP Tasks
    [Summarization, fill=violet!20
        [{En-Basic, En-Basic + \\ Variations, En-CoT, \\ Native-Basic, \\ Native-Basic  + Variations, \\ Native-CoT, Translate-En,\\ Translate-En-CoT, \\ X-Basic, XLT}, fill=violet!20
        [{
        Amharic, Arabic, Azerbaijani, Bengali, Burmese,\\ Chinese, Chinese (Traditional), English, French, Gaelic,\\ Gujarati, Hausa, Hindi, Igbo, Indonesian, German,\\ Japanese, Kirundi, Korean, Kyrgyz, Marathi, Nepali,\\ Oromo, Pashto, Persian, Pidgin, Portuguese, Punjabi,\\ Russian, Scottish Gaelic, Serbian, Serbian Cyrillic,\\ Serbian Latin, Sinhala, Somali, Spanish, Swahili, Tamil,\\ Telugu, Thai, Tigrinya, Turkish, Ukrainian, Urdu,\\ Uzbek, Vietnamese, Welsh, Yoruba  [\cite{huang2023not}\\, \cite{liu2024translation}, \cite{lai2023chatgpt}, \cite{asai2023buffet},\\ \cite{intrator2024breaking}, \cite{zhang2023cross},\\ \cite{ahuja2023mega}]
        }, fill=violet!20]
        ]
    ]
    [Task Understanding \\Consistency, fill=orange!20
        [En-Basic, fill=orange!20
        [{
        English-German, English-Chinese, Chinese-English, \\German-English [\cite{ohmer2023evaluating}]
        }, fill=orange!20]
        ]
    ]
    [Dialogue Evaluation, fill=green!20
        [{En-Basic, HD Prompting, \\MEEP, MEEP + Variations, \\G-Eval, GPTScore}, fill=green!20
        [{
        English, Spanish, \\Chinese  [\cite{ferron2023meep}, \cite{mendoncca2023simple}]
        }, fill=green!20]
        ]
    ]
    [Code-Mixing Natural\\ Language Inference, fill=white!20
        [{Native-Basic}, fill=white!20
        [{
        English-Hindi [\cite{ahuja2023mega}]
        }, fill=white!20]
        ]
    ]
    [Natural Language \\Inference, fill=yellow!20
        [{Native-CoT, \\ Translate-En-CoT, \\En-Basic, En-CoT, \\Translate-En, XLT, CLP, \\X-Basic, Native-Basic}, fill=yellow!20
        [{
        Arabic, Asháninka, Assamese, Aymara, Bengali, Bribri,\\ Bulgarian, Chinese, English, French, German, Greek,\\ Guarani, Gujarati, Hindi, Kannada, Korean, Malayalam,\\ Marathi, Modern Greek, Nahuatl, Oriya, Otomí,\\ Persian, Punjabi, Quechua, Rarámuri, Russian, \\Shipibo-Konibo, Spanish, Swahili, Tamil, Telugu,\\ Thai, Turkish, Urdu, Vietnamese, Wixarika [\\\cite{etxaniz2023multilingual}, \cite{huang2023not}, \\\cite{qin2023cross}, \cite{kim2023cross}, \\\cite{liu2024translation}, \cite{lai2023chatgpt}, \\\cite{asai2023buffet}, \cite{ahuja2023mega}]
        }, fill=yellow!20]
        ]
    ]
    [Relation Extraction, fill=pink!20
        [{En-Basic, ChatIE}, fill=pink!20
        [{
        English, Chinese, Russian, German, French, \\Spanish, Italian, Dutch, Polish, Portuguese, Arabic, \\Persian, Korean, Swedish, Ukrainian [ \\\cite{wei2023chatie}, \cite{lai2023chatgpt}]
        }, fill=pink!20]
        ]
    ]
    [Co-Reference \\ Resolution, fill=blue!20
        [{X-Basic, Native-Basic, \\En-Basic}, fill=blue!20
        [{
        Japanese, Portuguese, Russian, Chinese [\\\cite{asai2023buffet}]
        }, fill=blue!20]
        ]
    ]
    [Word Sense \\ Disambiguation, fill=white!20
        [{Native-Basic,  En-CoT}, fill=white!20
        [{
        Bulgarian, Danish, German, Estonian, Persian, French, \\Croatian, Italian, Japanese, Korean, Dutch, \\Chinese [\cite{shi2022language}]
        }, fill=white!20]
        ]
    ]
    [Event Extraction, fill=yellow!20
        [{En-Basic, ChatIE}, fill=yellow!20
        [{
        Chinese, English [\cite{wei2023chatie}]
        }, fill=yellow!20]
        ]
    ]
    [Named Entity \\ Recognition, fill=green!20
        [{En-Basic, ChatIE, \\X-Basic, Native-Basic}, fill=green!20
        [{
        Afrikaans, Amharic, Arabic, Assamese, Azerbaijani, \\Basque, Bengali, Belarusian, Bulgarian, Burmese, \\Chinese, Dutch, English, Estonian, Finnish, French, \\ Georgian, German, Greek, Guarani, Gujarati, Hausa, \\ Hebrew, Hindi, Hungarian, Igbo, Indonesian, Italian,\\ Japanese, Javanese, Kannada, Kazakh, Kinyarwanda,\\ Korean, Lithuanian, Luo, Malay, Malayalam, Marathi,\\ Modern Greek, NigerianPidgin, Oriya, Punjabi, Persian,\\ Polish, Portuguese, Punjabi, Quechua, Romanian,\\ Russian, Spanish, Swahili, Tagalog, Tamil, Telugu,\\ Thai, Turkish, Ukrainian, Urdu, Vietnamese, Wolof,\\ Yoruba
 [\cite{adelani2021masakhaner}, \cite{wei2023chatie}, \\\cite{asai2023buffet}, \cite{ahuja2023mega}]
        }, fill=green!20]
        ]
    ]
    [Part-Of-Speech \\ Tagging, fill=cyan!20
        [{En-Basic, Iterative \\Prompting, Decomposed \\Prompting (Generative), \\Decomposed Prompting \\(Probabilistic), Native-\\ Basic }, fill=cyan!20
        [{
        Afrikaans, Arabic, Basque, Bulgarian, Chinese,\\ Dutch, English, Estonian, Finnish, French, German,\\ Greek, Hebrew, Hindi, Hungarian, Indonesian,\\ Italian, Japanese, Kazakh, Korean, Lithuanian,\\ Mandarin, Marathi, Persian, Polish, Portuguese,\\ Romanian, Russian, Spanish, Tagalog, Tamil,\\ Telugu, Thai, Turkish, Ukrainian, Urdu, Vietnamese,\\ Wolof, Yoruba
 [\cite{lai2023chatgpt}, \cite{nie2024decomposed}, \\\cite{ahuja2023mega}]
        }, fill=cyan!20]
        ]
    ]
    ]
\end{forest}

\normalsize
\subsection{Mathematical Problem Solving}

\label{math}

The Mathematical Problem Solving task in NLP evaluates a model’s ability to comprehend, reason through and generate accurate solutions to mathematical problems presented in natural language. Unlike tasks that

\footnotesize
\begin{longtable}{p{.6in}p{.9in}p{1.4in}p{.5in}p{.8in}p{.8in}}
    \caption{Prompt Engineering Analysis for Mathematical Problem Solving Task}
    \label{tab:math} \\
    \toprule
    \textbf{Dataset} & \textbf{Prompting Strategy} & \textbf{LLM(s)}  & \textbf{SoTA} & \textbf{\#Languages} & \textbf{References} \\
    \midrule
    \endfirsthead
    
    \multicolumn{6}{c}%
    {{ \tablename\ \thetable{} Continued from the Previous Page}} \\
    \midrule
    \textbf{Dataset} & \textbf{Prompting Strategy} & \textbf{LLM(s)}  & \textbf{SoTA} & \textbf{\#Languages} & \textbf{References} \\
    \midrule
    \endhead

    \multicolumn{6}{c}{{\tablename\ \thetable{} Continued on the Next Page}} \\
    \endfoot

    \bottomrule
    \endlastfoot

    MGSM   & Native-Basic, Native-CoT, En-CoT, Translate-En-CoT, En-Basic, Translate-En, XLT, CLP, CLSP   & GPT-3 (Text-Davinci-002), PaLM-540B, XGLM-564M, XGLM-1.7B, XGLM-2.9B, XGLM-7.5B, LLaMA-7B, LLaMA-13B, LLaMA-30B, BLOOM-0.6B, BLOOM-1.7B, BLOOM-3B, BLOOM-7.1B, LLaMA-2-7B, LLaMA-2-13B, RedPajama-3B, RedPajama-7B, OpenLLaMA-3B, OpenLLaMA-7B, OpenLLaMA-13B, OpenLLaMA-V2-3B, OpenLLaMA-V2-7B, PolyLM-1B, PolyLM-13B, GPT-3.5 ( Text-Davinci-003), GPT-3.5-Turbo, LLaMA-2-70B-Chat, Codex (Code-Davinci-002), BLOOMZ-7.1B, LLaMA-2-13B, mT0-13B, Mistral-7B-Instruct, LLaMA-2-13B-Chat       &  CLSP & 11 & \cite{shi2022language, etxaniz2023multilingual, huang2023not, qin2023cross, liu2024translation} \\

\end{longtable}

\normalsize

rely solely on linguistic understanding, this task integrates language comprehension with symbolic and numerical reasoning. It requires models to interpret complex mathematical expressions, manipulate equations, perform multi-step arithmetic or algebraic computations and generate coherent, often step-by-step solutions. In multilingual settings, the task becomes even more challenging due to cross-linguistic variations in numerical formatting, mathematical terminology and syntactic structures. As such, it provides a robust benchmark for assessing both the reasoning abilities and cross-lingual generalization of LLMs. Despite the growing interest in multilingual evaluation, our survey identified only one relevant dataset for this task: MGSM \cite{shi2022language}, a multilingual extension of the GSM8K benchmark. Table \ref{tab:math} presents the dataset, the prompting methods explored, the number of languages covered across all strategies, the corresponding references and the most effective (SoTA) prompting technique reported in the literature.

\normalsize

\subsection{Commonsense Reasoning}

\label{commonsense}

Commonsense Reasoning is a core task in NLP that assesses a model’s ability to make inferences based on implicit, everyday human knowledge. Unlike tasks centered on factual retrieval, explicit logical reasoning or problem-solving, Commonsense Reasoning task emphasizes on a model’s capacity to understand and apply general world knowledge such as physical laws, social norms, temporal dynamics and causal relationships to interpret scenarios and make plausible judgments. In multilingual contexts, the task becomes even more complex due to linguistic and cultural differences in how commonsense knowledge is conveyed. Models must not only demonstrate robust reasoning skills but also adapt to the subtle and diverse ways this knowledge is encoded across languages. In our survey, we identified two prominent multilingual datasets for this task: XStoryCloze \cite{lin2022few}, which evaluates narrative understanding and X-CSQA \cite{lin2021common}, a cross-lingual extension of the CommonsenseQA benchmark. Table \ref{tab:commonr} presents these datasets alongside the prompting methods explored, the number of languages tested across all methods, the corresponding research works and the most effective prompting strategies identified to date.

\footnotesize
\begin{longtable}{p{.6in}p{.9in}p{1.4in}p{.5in}p{.8in}p{.8in}}
    \caption{Prompt Engineering Analysis for Commonsense Reasoning Task}
    \label{tab:commonr} \\
    \toprule
    \textbf{Dataset} & \textbf{Prompting Strategy} & \textbf{LLM(s)}  & \textbf{SoTA} & \textbf{\#Languages} & \textbf{References} \\
    \midrule
    \endfirsthead
    
    \multicolumn{6}{c}%
    {{ \tablename\ \thetable{} Continued from the Previous Page}} \\
    \midrule
    \textbf{Dataset} & \textbf{Prompting Strategy} & \textbf{LLM(s)}  & \textbf{SoTA} & \textbf{\#Languages} & \textbf{References} \\
    \midrule
    \endhead

    \multicolumn{6}{c}{{ \tablename\ \thetable{} Continued on the Next Page}} \\
    \endfoot
    
    \bottomrule
    \endlastfoot

    XStoryCloze  & Native-CoT, Translate-En-CoT, Native-Basic, Translate-En, X-Basic    & XGLM-564M, XGLM-1.7B, XGLM-2.9B, XGLM-7.5B, LLaMA-7B, LLaMA-13B, LLaMA-30B, BLOOM-0.6B, BLOOM-1.7B, BLOOM-3B, BLOOM-7.1B, LLaMA-2-7B, LLaMA-2-13B, RedPajama-3B, RedPajama-7B, OpenLLaMA-3B, OpenLLaMA-7B, OpenLLaMA-13B, OpenLLaMA-V2-3B, OpenLLaMA-V2-7B, PolyLM-1B, PolyLM-13B, PaLM 2-S, PaLM 2-L, GPT-3.5-Turbo, GPT-4, GPT-3.5 (Text-Davinci-003), XGLM, BLOOMZ      & Native-Basic & 11 & \cite{etxaniz2023multilingual, intrator2024breaking, ahuja2023mega} \\
    \midrule
    X-CSQA  & En-Basic    & GPT-3.5-Turbo    & En-Basic & 5 & \cite{lai2023chatgpt} \\

\end{longtable}

\footnotesize
\begin{longtable}{p{.6in}p{.9in}p{1.4in}p{.5in}p{.8in}p{.8in}}
    \caption{Prompt Engineering Analysis for Causal Reasoning Task}
    \label{tab:causalr} \\
    \toprule
    \textbf{Dataset} & \textbf{Prompting Strategy} & \textbf{LLM(s)}  & \textbf{SoTA} & \textbf{\#Languages} & \textbf{References} \\
    \midrule
    \endfirsthead
    
    \multicolumn{6}{c}%
    {{\bfseries \tablename\ \thetable{} -- continued from previous page}} \\
    \toprule
    \textbf{Dataset} & \textbf{Prompting Strategy} & \textbf{LLM(s)}  & \textbf{SoTA} & \textbf{\#Languages} & \textbf{References} \\
    \midrule
    \endhead

    \midrule \multicolumn{6}{c}{{Continued on next page}} \\
    \endfoot
    
    \bottomrule
    \endlastfoot

    XCOPA   & Native-Basic,  En-CoT, Native-CoT, Translate-En-CoT, En-Basic, Translate-En, XLT, CLP, CLSP, X-Basic      & Codex (Code-Davinci-002), PaLM-540B, XGLM-564M, XGLM-1.7B, XGLM-2.9B, XGLM-7.5B, LLaMA-7B, LLaMA-13B, LLaMA-30B, BLOOM-0.6B, BLOOM-1.7B, BLOOM-3B, BLOOM-7.1B, LLaMA-2-7B, LLaMA-2-13B, RedPajama-3B, RedPajama-7B, OpenLLaMA-3B, OpenLLaMA-7B, OpenLLaMA-13B, OpenLLaMA-V2-3B, OpenLLaMA-V2-7B, PolyLM-1B, PolyLM-13B, GPT-3.5 ( Text-Davinci-003), GPT-3.5-Turbo, GPT-4, LLaMA-2-70B-Chat, GPT-3 (Text-Davinci-002), BLOOMZ-7.1B, mT0-13B, mT5-13B, Mistral-7B-Instruct, LLaMA-2-13B-Chat, PaLM 2-S, PaLM 2-L, BLOOMZ, XGLM    & CLSP, Native-Basic & 15 & \cite{shi2022language, etxaniz2023multilingual, huang2023not, qin2023cross, kim2023cross, liu2024translation, asai2023buffet, intrator2024breaking, ahuja2023mega} \\

\end{longtable}

\normalsize
\subsection{Causal Reasoning}

Causal Reasoning is a critical task in NLP that assesses a model’s ability to identify, infer and reason about cause-and-effect relationships expressed in natural language. Unlike Commonsense Reasoning task, which draws on broad, implicit world knowledge, Causal Reasoning task requires a more structured understanding of how events or concepts influence one another. This includes recognizing both explicit and implicit causal cues, distinguishing causation from correlation and generating or selecting plausible causes or effects based on a given context. 
In multilingual settings, Causal Reasoning task presents additional challenges due to cross-linguistic variations in the expression of causality such as differences in causal connectives, syntactic structures  and culturally embedded reasoning patterns. LLMs must generalize their understanding of causality across languages, often under low-resource conditions. For this task, we found one widely adopted multilingual dataset: XCOPA \cite{ponti2020xcopa}, a multilingual adaptation of the Choice of Plausible Alternatives (COPA) benchmark. Table \ref{tab:causalr} lists the dataset, the prompting methods employed, the total number of unique languages tested, citations of relevant work and the SoTA prompting approach identified in multilingual settings.

\normalsize
\subsection{Contextual Question-Answering}
Contextual Question Answering task tests a model’s ability to comprehend and extract or infer answers to questions based solely on a given textual context. Unlike open-domain or knowledge-based question-answering, where models may draw upon external sources or background knowledge, Contextual Question Answering restricts the model to rely entirely on the provided passage, paragraph or document to answer the question accurately. In multilingual settings, Contextual Question Answering task introduces additional complexity, as models must handle variations in syntax, semantics and discourse structures across different languages while preserving contextual alignment between the question and the answer. In our survey, we examined four key datasets frequently used in multilingual Contextual Question Answering NLP task: MLQA \cite{lewis2019mlqa}, XQuAD \cite{artetxe2019cross}, TyDiQA-GP \cite{clark2020tydi}, IndicQA \cite{doddapaneni2022indicxtreme}, and Belebele \cite{bandarkar2023belebele}. Table \ref{tab:cqa} lists these datasets along with the prompting methods used, the total number of unique languages covered across all methods, citations for associated research papers and the identified SoTA prompting strategies.

\footnotesize
\begin{longtable}{p{.6in}p{.9in}p{1.4in}p{.5in}p{.8in}p{.8in}}
    \caption{Prompt Engineering Analysis for Contextual Question-Answering Task}
    \label{tab:cqa} \\
    \toprule
    \textbf{Dataset} & \textbf{Prompting Strategy} & \textbf{LLM(s)}  & \textbf{SoTA} & \textbf{\#Languages} & \textbf{References} \\
    \midrule
    \endfirsthead
    
    \multicolumn{6}{c}%
    {{ \tablename\ \thetable{} Continued from the Previous Page}} \\
    \toprule
    \textbf{Dataset} & \textbf{Prompting Strategy} & \textbf{LLM(s)}  & \textbf{SoTA} & \textbf{\#Languages} & \textbf{References} \\
    \midrule
    \endhead

    \multicolumn{6}{c}{{ \tablename\ \thetable{} Continued on the Next Page}} \\
    \endfoot
    
    \bottomrule
    \endlastfoot

    MLQA   & En-Basic, Native-Basic, X-Basic, Translate-En    & XGLM-1.7B, XGLM-2.9B, XGLM-7.5B, BLOOM-1.1B, BLOOM-3B, BLOOM-7.1B, GPT-3.5-Turbo, GPT-4, GPT-3.5 (Text-Davinci-003), XGLM, BLOOMZ    & Native-Basic & 7 & \cite{kim2023cross, ahuja2023mega} \\
        \midrule
        XQuAD   & Native-Basic, Translate-En, X-Basic, En-Basic, SAP      & XGLM-1.7B, XGLM-2.9B, XGLM-7.5B, BLOOM-1.1B, BLOOM-3B, BLOOM-7.1B, mT5-3.7B, GPT-3.5-Turbo, GPT-4, GPT-3.5 (Text-Davinci-003), XGLM, BLOOMZ, PaLM 2-S, PaLM 2-L    & Native-Basic & 12 & \cite{kim2023cross, patel2022bidirectional, lai2023chatgpt, intrator2024breaking, ahuja2023mega} \\
        \midrule
        TyDiQA-GP   & X-Basic, En-Basic, Native-Basic, Translate-En  &  BLOOMZ-7.1B, BLOOM-7.1B, GPT-3.5-Turbo, GPT-4, GPT-3.5 (Text-Davinci-003), XGLM, BLOOMZ,  mT5-13B, mT0-13B, PaLM 2-S, PaLM 2-L & Native-Basic & 9  & \cite{asai2023buffet, intrator2024breaking, ahuja2023mega} \\
        \midrule
        BeleBele   & Native-Basic, Translate-En   & PaLM 2-S, PaLM 2-L    & Native-Basic & 104 & \cite{intrator2024breaking}  \\
        \midrule
        IndicQA   & X-Basic   & GPT-3.5-Turbo, GPT-4, GPT-3.5 (Text-Davinci-003), XGLM, BLOOMZ   & X-Basic & 11 & \cite{ahuja2023mega}  \\
    
\end{longtable}

\normalsize

\subsection{Context-Free Question-Answering}

Context-Free Question-Answering refers to the task of answering questions without access to a task-specific context or passage, relying instead on a model’s internal knowledge or external open-domain resources (e.g., Wikipedia or the web). Unlike Contextual Question-Answering task, where answers are derived from a given reference text, Context-Free Question-Answering challenges the model to retrieve, recall or infer factual information purely from pre-trained knowledge or retrieval-based augmentation. This task emphasizes a model's ability to generalize across diverse knowledge domains and languages, and requires robust information recall, reasoning and disambiguation skills. Multilingual Context-Free Question-Answering further complicates this task by introducing linguistic diversity in how questions are phrased, as well as potential cultural variations in knowledge representation. Moreover, since models must handle diverse question styles and entity references across languages without specific contextual grounding, effective prompting and knowledge alignment strategies are essential for achieving strong performance. In our work, we came across two multilingual datasets relevant to this task: MKQA \cite{longpre2021mkqa}, which offers manually curated, language-diverse question-answering pairs aligned with English queries and Wikipedia Cloze-QA \\\cite{kakwani2020indicnlpsuite}, a cloze-style question-answering dataset built using Wikipedia articles in multiple Indic languages. Table \ref{tab:cfqa} outlines these datasets, the prompting methods applied to them, the cumulative number of languages covered, associated research studies and the best-performing prompting techniques.

\footnotesize
\begin{longtable}{p{.6in}p{.9in}p{1.4in}p{.5in}p{.8in}p{.8in}}
    \caption{Prompt Engineering Analysis for Context-Free Question-Answering Task}
    \label{tab:cfqa} \\
    \toprule
    \textbf{Dataset} & \textbf{Prompting Strategy} & \textbf{LLM(s)}  & \textbf{SoTA} & \textbf{\#Languages} & \textbf{References} \\
    \midrule
    \endfirsthead
    
    \multicolumn{6}{c}%
    {{ \tablename\ \thetable{} Continued from the Previous Page}} \\
    \midrule
    \textbf{Dataset} & \textbf{Prompting Strategy} & \textbf{LLM(s)}  & \textbf{SoTA} & \textbf{\#Languages} & \textbf{References} \\
    \midrule
    \endhead

    \multicolumn{6}{c}{{ \tablename\ \thetable{} Continued on the Next Page}} \\
    \endfoot
    
    \bottomrule
    \endlastfoot

    MKQA  & En-Basic, En-CoT, XLT, Native-Basic, Native-CoT, Translate-En-CoT   & GPT-3.5 ( Text-Davinci-003), GPT-3.5-Turbo, LLaMA-2-70B-Chat, Codex (Code-Davinci-002), Mistral-7B-Instruct, LLaMA-2-13B-Chat, BLOOMZ-7.1B      & XLT & 10 & \cite{huang2023not, liu2024translation} \\
        \midrule
        Wikipedia Cloze-QA  & En-Basic  & GPT-4, GPT-3.5-Turbo      & En-Basic & 11 & \cite{lai2023chatgpt} \\
  
\end{longtable}

\normalsize
\subsection{Subjectivity Analysis}

Subjectivity Analysis task determines whether a given text expresses subjective information, such as opinions, feelings or sentiments, as opposed to objective facts. It aims to identify the subjective or emotional stance of the author, categorizing text as either subjective (expressing personal views or feelings) or objective (conveying factual information). The task involves several challenges, such as distinguishing between subjective and objective language in texts with nuanced meanings, sarcasm or implied sentiment. In multilingual settings, the complexity is compounded by linguistic and cultural variations in how subjectivity is expressed. A model needs to understand not only the specific language but also the broader cultural context in which the sentiment or opinion is conveyed. While traditional methods focused on feature engineering and classifiers like SVMs or logistic regression, modern approaches leverage deep learning techniques, particularly recurrent neural networks (RNNs) and transformers, to better capture the subtlety of subjective language. Furthermore, pre-trained multilingual models like mBERT and XLM-R are being increasingly used for subjectivity analysis tasks, as they can capture shared linguistic representations across multiple languages. In the course of our survey, we identified one dataset relevant to Subjectivity Analysis which is X-Sub \cite{dwivedi2024navigating}. Table \ref{tab:suba} presents the identified dataset, the prompting methods explored, the number of languages covered, relevant research studies and the SoTA prompting technique. 

\normalsize
\subsection{Emotion/Sentiment Understanding}

This task focuses on determining the emotional tone or sentiment expressed in a given text. This task involves identifying whether a text conveys a positive, negative or neutral sentiment, as well as understanding more

\footnotesize
\begin{longtable}{p{.6in}p{.9in}p{1.4in}p{.5in}p{.8in}p{.8in}}
    \caption{Prompt Engineering Analysis for Subjectivity Analysis Task}
    \label{tab:suba} \\
    \toprule
    \textbf{Dataset} & \textbf{Prompting Strategy} & \textbf{LLM(s)}  & \textbf{SoTA} & \textbf{\#Languages} & \textbf{References} \\
    \midrule
    \endfirsthead
    
    \multicolumn{6}{c}%
    {{\bfseries \tablename\ \thetable{} -- continued from previous page}} \\
    \toprule
    \textbf{Dataset} & \textbf{Prompting Strategy} & \textbf{LLM(s)}  & \textbf{SoTA} & \textbf{\#Languages} & \textbf{References} \\
    \midrule
    \endhead

    \midrule \multicolumn{6}{c}{{Continued on next page}} \\
    \endfoot
    
    \bottomrule
    \endlastfoot

    X-Sub & Regressive En-CoT, Regressive Native-CoT     & GPT-4, BARD     & Regressive Native-CoT & 9 & \cite{dwivedi2024navigating}\\

\end{longtable}

\normalsize
complex emotional states such as happiness, anger, sadness, fear, or surprise. Emotion/Sentiment Understanding plays a vital role in numerous applications, including social media monitoring, customer service chatbots and market research. In multilingual contexts, Emotion/Sentiment Understanding becomes more challenging due to variations in how emotions are expressed across languages and cultures. The nuances of sentiment can vary significantly, making it crucial for models to account for these differences when processing texts in multiple languages. Additionally, the availability of labeled datasets in different languages remains a key challenge, as emotions and sentiment can be expressed differently across cultural and linguistic boundaries. Our survey identified three multilingual datasets that have been used for Emotion/Sentiment Understanding NLP task. These datasets include MARC \cite{keung2020multilingual}, CLS \cite{prettenhofer2010cross} and IndicNLU-Sentiment \cite{aggarwal2022indicxnli}. Table \ref{tab:emotion} lists these datasets, the prompting methods applied, the cumulative number of languages experimented with and the SoTA prompting strategies identified through experimental results.

\normalsize
\subsection{Code-Mixing Emotion/Sentiment Understanding}

Code-Mixing Emotion/Sentiment Understanding is a task in multilingual NLP that focuses on identifying and interpreting emotional tone or sentiment polarity in texts that contain code-mixed language i.e., utterances that alternate between two or more languages within the same sentence or discourse. Unlike traditional sentiment analysis, this task poses additional challenges due to irregular syntax, mixed grammatical structures, transliterations and varied usage of idiomatic expressions across languages. The complexity of language-switching demands models capable of capturing nuanced sentiment expressions that may not be fully conveyed in monolingual contexts. From a computational perspective, this task requires models to handle lexical and syntactic irregularities, recognize emotion cues spread across multiple languages and generalize well across diverse code-mixing patterns. Moreover, emotion and sentiment expressions in code-mixed text are often culturally contextual and informal, making them difficult for traditional language models to process without task-specific adaptation. In the context of our survey, we found LinCE \cite{aguilar2020lince}, MixSentiment Malayalam \cite{chakravarthi2020sentiment}, EN-ES-CS \cite{vilares2016cs} and MixSentiment Tamil \cite{chakravarthi2020corpus} datasets. Table \ref{tab:cmemotion} details the prompting strategies applied, the number of language combinations explored, the relevant studies and the best-performing prompting method observed.

\normalsize
\subsection{Toxicity Understanding}

Toxicity Understanding aims to identify and classify toxic language, which includes harmful, offensive or inappropriate content. Toxicity can manifest in various forms, such as hate speech, harassment, threats, discriminatory remarks or any language that can potentially harm or distress individuals or groups. The task requires models to detect such language within text, even when it is implicit, indirect or subtle. In multilingual settings, detecting toxicity becomes even more complex due to linguistic and cultural differences in how toxic expressions are framed and perceived. The language structure and tone may vary and expressions of toxicity in one language may not directly translate into the same form in another language. To address these challenges, it is crucial to develop cross-lingual toxicity detection models that can generalize across languages and cultures, while being sensitive to local nuances in toxic speech. In our presented work, we came across only two datasets which are HatEval \cite{basile2019semeval} and \href{ https://www.kaggle.com/competitions/jigsaw-multilingual-toxic-comment-classification}{Jigsaw}. Table \ref{tab:toxic} presents these datasets, the various prompting techniques explored, the number of languages covered by these datasets, citations of relevant studies and the most effective (SoTA) prompting method identified in multilingual settings.

\footnotesize
\begin{longtable}{p{.6in}p{.9in}p{1.4in}p{.5in}p{.8in}p{.8in}}
    \caption{Prompt Engineering Analysis for Emotion/Sentiment Understanding Task}
    \label{tab:emotion} \\
    \toprule
    \textbf{Dataset} & \textbf{Prompting Strategy} & \textbf{LLM(s)}  & \textbf{SoTA} & \textbf{\#Languages} & \textbf{References} \\
    \midrule
    \endfirsthead
    
    \multicolumn{6}{c}%
    {{ \tablename\ \thetable{} Continued from the Previous Page}} \\
    \toprule
    \textbf{Dataset} & \textbf{Prompting Strategy} & \textbf{LLM(s)}  & \textbf{SoTA} & \textbf{\#Languages} & \textbf{References} \\
    \midrule
    \endhead

    \multicolumn{6}{c}{{ \tablename\ \thetable{} Continued on the Next Page}}
    \endfoot
    
    \bottomrule
    \endlastfoot

    MARC & Random Prompting, Semantic Alignment, Task-Based Alignment, X-InSTA, X-Basic, Native-Basic, En-Basic    & XGLM-7.5B, BLOOMZ-7.1B, BLOOM-7.1B, GPT-3.5-Turbo, mT5-13B, mT0-13B   & X-InSTA  & 6 & \cite{tanwar2023multilingual, asai2023buffet} \\
        \midrule
        CLS & Random Prompting, Semantic Alignment, Task-Based Alignment, X-InSTA       & XGLM-7.5B     & X-InSTA & 4 & \cite{tanwar2023multilingual} \\
        \midrule
         IndicNLU-Sentiment  &  X-Basic, Native-Basic, En-Basic     & BLOOMZ-7.1B, BLOOM-7.1B, GPT-3.5-Turbo, mT5-13B, mT0-13B    & En-Basic & 14 & \cite{asai2023buffet} \\

\end{longtable}

\footnotesize
\begin{longtable}{p{.6in}p{.9in}p{1.4in}p{.5in}p{.8in}p{.8in}}
    \caption{Prompt Engineering Analysis for Code-Mixing Emotion/Sentiment Understanding Task}
    \label{tab:cmemotion} \\
    \toprule
    \textbf{Dataset} & \textbf{Prompting Strategy} & \textbf{LLM(s)}  & \textbf{SoTA} & \textbf{\#Language Pairs} & \textbf{References} \\
    \midrule
    \endfirsthead
    
    \multicolumn{6}{c}%
    {{ \tablename\ \thetable{} Continued from the Previous Page}} \\
    \toprule
    \textbf{Dataset} & \textbf{Prompting Strategy} & \textbf{LLM(s)}  & \textbf{SoTA} & \textbf{\#Language Pairs} & \textbf{References} \\
    \midrule
    \endhead

    \multicolumn{6}{c}{{\tablename\ \thetable{} Continued on the Next Page}} \\
    \endfoot
    
    \bottomrule
    \endlastfoot

    LinCE & En-Basic, En-Basic + Variations     & GPT-3.5-Turbo, BLOOMZ-0.6B, BLOOMZ-1.1B, BLOOMZ-1.7B, BLOOMZ-3B, BLOOMZ-7.1B, mT0-300M, mT0-580M, mT0-1.2B, mT0-3.7B, mT0-13B, XGLM-564M, XGLM-1.7B, XGLM-2.9B, XGLM-4.5B, XGLM-7.5B     & En-Basic + Variations & 1 & \cite{zhang2023multilingual} \\
        \midrule
        EN-ES-CS   &  Native-Basic     & GPT-3.5-Turbo, GPT-4, GPT-3.5 (Text-Davinci-003), XGLM, BLOOMZ    & Native-Basic & 1 & \cite{ahuja2023mega} \\
        \midrule
        Mix-Sentiment Malayalam & En-Basic, En-Basic + Variations      & GPT-3.5-Turbo, BLOOMZ-0.6B, BLOOMZ-1.1B, BLOOMZ-1.7B, BLOOMZ-3B, BLOOMZ-7.1B, mT0-300M, mT0-580M, mT0-1.2B, mT0-3.7B, mT0-13B, XGLM-564M, XGLM-1.7B, XGLM-2.9B, XGLM-4.5B, XGLM-7.5B      & En-Basic + Variations & 1 & \cite{zhang2023multilingual} \\
        \midrule
         Mix-Sentiment Tamil   &  En-Basic, En-Basic + Variations     & GPT-3.5-Turbo, BLOOMZ-0.6B, BLOOMZ-1.1B, BLOOMZ-1.7B, BLOOMZ-3B, BLOOMZ-7.1B, mT0-300M, mT0-580M, mT0-1.2B, mT0-3.7B, mT0-13B, XGLM-564M, XGLM-1.7B, XGLM-2.9B, XGLM-4.5B, XGLM-7.5B    & En-Basic + Variations & 1 & \cite{zhang2023multilingual} \\

\end{longtable}

\footnotesize
\begin{longtable}{p{.6in}p{.9in}p{1.4in}p{.5in}p{.8in}p{.8in}}
    \caption{Prompt Engineering Analysis for Toxicity Understanding Task}
    \label{tab:toxic} \\
    \toprule
    \textbf{Dataset} & \textbf{Prompting Strategy} & \textbf{LLM(s)}  & \textbf{SoTA} & \textbf{\#Languages} & \textbf{References} \\
    \midrule
    \endfirsthead
    
    \multicolumn{6}{c}%
    {{\bfseries \tablename\ \thetable{} -- continued from previous page}} \\
    \toprule
    \textbf{Dataset} & \textbf{Prompting Strategy} & \textbf{LLM(s)}  & \textbf{SoTA} & \textbf{\#Languages} & \textbf{References} \\
    \midrule
    \endhead

    \midrule \multicolumn{6}{c}{{Continued on next page}} \\
    \endfoot
    
    \bottomrule
    \endlastfoot

    HatEval & Random Prompting, Semantic Alignment, Task-Based Alignment, X-InSTA   & XGLM-7.5B   & X-InSTA & 2 & \cite{tanwar2023multilingual}\\

    Jigsaw & Native-Basic, X-Basic, Translate-En   & GPT-3.5-Turbo, GPT-4, GPT-3.5 (Text-Davinci-003), XGLM, BLOOMZ & X-Basic & 6 & \cite{ahuja2023mega}\\

\end{longtable}

\normalsize
\subsection{Question Generation}

Question Generation is a task focused on the automatic creation of questions from a given input, which can be a passage of text, an image, or a structured dataset. The task involves not only understanding the content but also generating questions that are both syntactically well-formed and semantically relevant to the given context. In Question Generation, the model is tasked with identifying important pieces of information from the input and then generating a variety of questions, such as factual, reasoning, or descriptive questions, depending on the nature of the input content. Traditionally, Question Generation systems have relied on rule-based approaches, where templates were defined to generate questions from specific patterns in the text. However, recent advancements have utilized neural network-based models, particularly sequence-to-sequence architectures and pre-trained transformer models, such as BERT and T5. These models have significantly improved the quality and diversity of generated questions by learning to map between input text and question-like output representations. In the multilingual context, Question Generation becomes even more complex due to variations in question structures and linguistic features across languages. For example, some languages may require changes in word order, the use of gendered pronouns or specific question markers, making it necessary to adapt Question Generation models to handle these differences effectively. In our survey, we identified one multilingual dataset relevant to the Question Generation task which includes TyDiQA-GP-QG \cite{xiao2020ernie}. Table~\ref{tab:qg} lists this dataset, the different prompting methods applied, the number of languages tested, relevant studies, and the SoTA prompting technique observed across this dataset.

\footnotesize
\begin{longtable}{p{.6in}p{.9in}p{1.4in}p{.5in}p{.8in}p{.8in}}
    \caption{Prompt Engineering Analysis for Question Generation Task}
    \label{tab:qg} \\
    \toprule
    \textbf{Dataset} & \textbf{Prompting Strategy} & \textbf{LLM(s)}  & \textbf{SoTA} & \textbf{\#Languages} & \textbf{References} \\
    \midrule
    \endfirsthead
    
    \multicolumn{6}{c}%
    {{\bfseries \tablename\ \thetable{} -- continued from previous page}} \\
    \toprule
    \textbf{Dataset} & \textbf{Prompting Strategy} & \textbf{LLM(s)}  & \textbf{SoTA} & \textbf{\#Languages} & \textbf{References} \\
    \midrule
    \endhead

    \midrule \multicolumn{6}{c}{{Continued on next page}} \\
    \endfoot
    
    \bottomrule
    \endlastfoot

    TyDiQA-GP-QG & X-Basic, Native-Basic, En-Basic     & BLOOMZ-7.1B, BLOOM-7.1B, GPT-3.5-Turbo, mT5-13B, mT0-13B      & X-Basic & 8 & \cite{asai2023buffet}\\

\end{longtable}

\normalsize
\subsection{Code-Mixing Generation}

Code-Mixing Generation is a specialized NLP task that focuses on the ability of models to generate text that contains code-mixed language, where elements from two or more languages are interwoven within a sentence or conversation. Code-mixing typically occurs in multilingual environments, where speakers alternate between languages or use multiple languages within the same utterance. This task involves not only understanding the syntax and semantics of each language involved but also mastering the seamless integration of linguistic structures from different languages in a coherent manner. The Code-Mixing task is particularly complex in the generation domain because the model must go beyond simple translation or substitution of words from one language into another. It must respect the social, cultural and syntactic nuances that govern code-switching behavior in natural conversation. Furthermore, challenges arise in handling language pairs that exhibit different levels of resource availability or structural differences. In our survey, we identified only one study in which no gold-standard dataset was employed; instead, the generated outputs were evaluated directly by human annotators. Table \ref{tab:cmgeneration} summarizes the prompting methods explored, the total number of unique language pairs covered in the experiments, the relevant studies and the SoTA prompting strategy identified.

\footnotesize
\begin{longtable}{p{.6in}p{.9in}p{1.4in}p{.5in}p{.8in}p{.8in}}
    \caption{Prompt Engineering Analysis for Code-Mixing Generation Task}
    \label{tab:cmgeneration} \\
    \toprule
    \textbf{Dataset} & \textbf{Prompting Strategy} & \textbf{LLM(s)}  & \textbf{SoTA} & \textbf{\#Language Pairs} & \textbf{References} \\
    \midrule
    \endfirsthead
    
    \multicolumn{6}{c}%
    {{\bfseries \tablename\ \thetable{} -- continued from previous page}} \\
    \toprule
    \textbf{Dataset} & \textbf{Prompting Strategy} & \textbf{LLM(s)}  & \textbf{SoTA} & \textbf{\#Language Pairs} & \textbf{References} \\
    \midrule
    \endhead

    \midrule \multicolumn{6}{c}{{Continued on next page}} \\
    \endfoot
    
    \bottomrule
    \endlastfoot

    -- & En-Basic, En-Basic + Variations    & GPT-3.5-Turbo, InstructGPT (Text-Davinci-003), InstructGPT (Text-Davinci-002), BLOOMZ-176B, Flan-T5-XXL    & En-Basic + Variations & 7 & \cite{yong2023prompting} \\

\end{longtable}

\normalsize
\subsection{Machine Translation}

Machine Translation is a foundational task in NLP that focuses on automatically translating text from one language to another while preserving semantic fidelity, fluency and cultural relevance. It serves as a core application of multilingual language modeling and remains a critical component in bridging linguistic divides, enabling cross-lingual information access and supporting global communication. Modern Machine Translation approaches have evolved from traditional rule-based and statistical paradigms to Neural Machine Translation (NMT) systems powered by Transformer architectures and large-scale pre-trained language models. These models aim to capture complex syntactic and semantic correspondences between source and target languages and have demonstrated remarkable performance in high-resource language pairs. However, the task becomes significantly more challenging in multilingual and low-resource settings, where issues such as vocabulary mismatch, data sparsity, typological divergence and domain shifts arise. Addressing these challenges requires innovations in prompt design, parameter-efficient fine-tuning, data augmentation and transfer learning. Recent studies also explore zero-shot and few-shot translation capabilities of LLMs, revealing their potential for handling previously unseen language pairs with minimal supervision. While extensive multilingual benchmarks exist, our survey focuses on a set of them including FLORES \cite{costa2022no}, FLORES-101 \cite{goyal2022flores}, MLIKS \cite{aharoni2020unsupervised},  AMBIGMT \cite{pilault2023interactive}, WMT-22 \cite{kocmi2022findings}, WMT 21 \cite{freitag2021experts}, WMT 20 \cite{freitag2021experts}, DiBiMT \cite{campolungo2022dibimt}, My Uncle Napoleon \cite{pourkamali2024machine}, The Beggar Boy at Christ’s Christmas Tree \cite{pourkamali2024machine}, The Tell-Tale Heart \cite{pourkamali2024machine}, TICO-19 \cite{anastasopoulos2020tico} and {\href{https://www.discovermagazine.com/the-sciences/a-major-time-travel-perk-may-be-technically-impossible}{A Major Time Travel Perk May Be Technically Impossible}}. Table~\ref{tab:mtrans} highlights the datasets involved in these prompting-based translation setups, the number of languages covered and the prompting strategies that demonstrated superior performance across various translation scenarios. Additionally, the table includes relevant research citations corresponding to various prompting methods, providing a comprehensive view of the empirical efforts supporting these techniques.

\footnotesize
\begin{longtable}{p{.6in}p{.9in}p{1.4in}p{.5in}p{.8in}p{.8in}}
    \caption{Prompt Engineering Analysis for Machine Translation Task}
    \label{tab:mtrans} \\
    \toprule
    \textbf{Dataset} & \textbf{Prompting Strategy} & \textbf{LLM(s)}  & \textbf{SoTA} & \textbf{\#Language Pairs} & \textbf{References} \\
    \midrule
    \endfirsthead
    
    \multicolumn{6}{c}%
    {{ \tablename\ \thetable{} Continued from the Previous Page}} \\
    \midrule
    \textbf{Dataset} & \textbf{Prompting Strategy} & \textbf{LLM(s)}  & \textbf{SoTA} & \textbf{\#Language Pairs} & \textbf{References} \\
    \midrule
    \endhead

    \multicolumn{6}{c}{{\tablename\ \thetable{} Continued on the Next Page}} \\
    \endfoot
    
    \bottomrule
    \endlastfoot

    FLORES & En-Basic, En-CoT, XLT, CoD, CoD + Variations, DecoMT, SAP, TALENT     & Alpaca-7B, BLOOM-176B, BLOOM-7B, BLOOM-7.1B, BLOOMZ-176B, BLOOMZ-7.1B, Codex (Code-Davinci-002), GPT-3.5 ( Text-Davinci-003), GPT-3.5-Turbo, InstructGPT (Text-Davinci-003), LLaMA-2-70B-Chat, LLaMA-65B, LLaMA-7B, XGLM-7.5B, mT5-XXL-3.7B
    & DecoMT & 406 & \cite{huang2023not, lu2023chain, puduppully2023decomposed, iyer2023towards, guo2024teaching}\\
    \midrule
    FLORES-101 & En-Basic, DIPMT, DIPMT + Variations, SAP, SAP + Variations     & OPT-175B, BLOOM-176B, XGLM-7.5B, mT5-3.7B, GPT-3-6.7B      & DIPMT + Variations & 200 & \cite{ghazvininejad2023dictionary, patel2022bidirectional} \\
        \midrule
        MLIKS & En-Basic, DIPMT, HIL, HIL + Variations     & OPT-175B, BLOOM-176B, GPT-3.5-Turbo      & DIPMT & 1 & \cite{ghazvininejad2023dictionary, yang2023human} \\
        \midrule
         AMBIGMT  & En-Basic, Ambiguity Context, InterCPT      & PaLM-540B    & InterCPT & 4 & \cite{pilault2023interactive} \\
        \midrule
         
         WMT-21  & Translator + Post Edit    & GPT-4, GPT-3.5-Turbo     & Translator + Post Edit & 1 & \cite{raunak2023leveraging} \\
         \midrule
         WMT-20  & Translator + Post Edit     & GPT-4, GPT-3.5-Turbo     & Translator + Post Edit & 1 & \cite{raunak2023leveraging} \\
         \midrule
         
         DiBiMT  & En-Basic    & LLaMA-7B, LLaMA-65B , Alpaca-7B, BLOOM-7.1B, BLOOM-176B, BLOOMZ-7.1B, BLOOMZ-176B     & En-Basic & 5 & \cite{iyer2023towards} \\
         \midrule

         WMT-22  & En-Basic, Rerank, MAPS, Translator + Post Edit, Translator + Post Edit CoT, Translator + Post Edit SCoT, Multi-Turn Dialogue, Multi-Turn Dialogue + Variations     & GPT-3.5 (Text-Davinci-003), Vicuna-7B, Alpaca-7B, GPT-4, GPT-3.5-Turbo     & MAPS & 11 & \cite{he2024exploring, raunak2023leveraging, wu2023exploring} \\
         \midrule
         
         My Uncle Napoleon  & En-Basic    & GPT-3.5-Turbo, GPT-4, PaLM 2, LLaMA-2-70B-Chat, Claude 2, Perplexity AI + CoPilot     & En-Basic & 2 & \cite{pourkamali2024machine} \\
         \midrule
         The Beggar Boy at Christ’s Christmas Tree  & En-Basic     & GPT-3.5-Turbo, GPT-4, PaLM 2, LLaMA-2-70B-Chat, Claude 2, Perplexity AI + CoPilot     & En-Basic & 2 & \cite{pourkamali2024machine} \\
         \midrule
         The Tell-Tale Heart  & En-Basic     & GPT-3.5-Turbo, GPT-4, PaLM 2, LLaMA-2-70B-Chat, Claude 2, Perplexity AI + CoPilot     & En-Basic & 2 & \cite{pourkamali2024machine} \\
         \midrule
         TICO-19  & En-Basic, Fuzzy Match, Translator + Fuzzy Match, Fuzzy Match + Glossary & GPT-3.5 (Text-Davinci-003), BLOOM, BLOOMZ & Translator + Fuzzy Match & 5 & \cite{moslem2023adaptive} \\
         \midrule
         A Major Time Travel Perk May Be Technically Impossible  & En-Basic, En-Basic + Variations     & GPT-4     & En-Basic + Variations & 1 & \cite{he2024prompting} \\

\end{longtable}

\normalsize
\subsection{Code-Mixing Machine Translation}

Code-Mixing Machine Translation is a challenging task in multilingual NLP that involves translating sentences or utterances that contain code-mixed language i.e., a mixture of two or more languages into a monolingual target language. Unlike standard Machine Translation, where source and target languages are clearly defined and linguistically coherent, code-mixed inputs often include informal expressions, non-standard grammar, and frequent switches between languages, making them difficult to process with conventional translation systems. This task is particularly relevant in multilingual societies where speakers commonly alternate between languages in speech and informal text (e.g., social media, chat messages). Effective translation in such settings requires models to understand context-sensitive language switches, normalize noisy or transliterated inputs and disambiguate meaning across the mixed-language span. Furthermore, the semantic representation of code-mixed inputs often does not align neatly with grammatical structures in the target language, necessitating advanced contextual understanding and adaptation mechanisms within the translation model. While reading up papers for our survey, we found one relevant dataset MixMT-2022 \cite{srivastava2022overview}. Table \ref{tab:cmtranslation} presents this dataset, the prompting methods explored, the number of language pairs involved, associated studies and the most effective prompting strategy identified in this domain for the discussed dataset.

\footnotesize
\begin{longtable}{p{.6in}p{.9in}p{1.4in}p{.5in}p{.8in}p{.8in}}
    \caption{Prompt Engineering Analysis for Code-Mixing Machine Translation Task}
    \label{tab:cmtranslation} \\
    \toprule
    \textbf{Dataset} & \textbf{Prompting Strategy} & \textbf{LLM(s)}  & \textbf{SoTA} & \textbf{\#Language Pairs} & \textbf{References} \\
    \midrule
    \endfirsthead
    
    \multicolumn{6}{c}%
    {{\bfseries \tablename\ \thetable{} -- continued from previous page}} \\
    \toprule
    \textbf{Dataset} & \textbf{Prompting Strategy} & \textbf{LLM(s)}  & \textbf{SoTA} & \textbf{\#Language Pairs} & \textbf{References} \\
    \midrule
    \endhead

    \midrule \multicolumn{6}{c}{{Continued on next page}} \\
    \endfoot
    
    \bottomrule
    \endlastfoot

    MixMT-2022 & En-Basic, En-Basic + Variations     & GPT-3.5-Turbo, BLOOMZ-560M, BLOOMZ-1.1B, BLOOMZ-1.7B, BLOOMZ-3B, BLOOMZ-7.1B, mT0-300M, mT0-580M, mT0-1.2B, mT0-3.7B, mT0-13B, XGLM-564M, XGLM-1.7B, XGLM-2.9B, XGLM-4.5B, XGLM-7.5B     & En-Basic + Variations & 1 & \cite{zhang2023multilingual} \\

\end{longtable}

\normalsize
\subsection{Code-Mixing Language Identification}

Code-Mixing Language Identification is a fundamental task in multilingual NLP, particularly relevant to informal communication settings such as social media, instant messaging and spoken dialogue. This task focuses on detecting the language of individual tokens or phrases within a code-mixed utterance, where words from two or more languages are interleaved within a single sentence or conversation turn. The goal is to assign a language tag to each token, enabling downstream applications like code-mixed speech recognition, sentiment analysis and translation to function effectively. Unlike monolingual language identification, which deals with texts in a single language, this task must handle abrupt language switches, orthographic variations, transliterations and overlapping vocabulary across languages. The presence of named entities, borrowed terms, and informal spelling further complicates the identification process. Our review of the literature revealed one prominent dataset LinCE \cite{aguilar2020lince}. Table \ref{tab:cmlid} highlights the dataset, the prompting techniques applied, the number of language pairs involved, the relevant research contributions and the most effective prompting strategy observed in this domain.

\normalsize
\subsection{Part-Of-Speech Tagging}

Part-Of-Speech Tagging is a sequence labeling task in NLP that involves assigning grammatical categories such as noun, verb, adjective or preposition—to each word in a sentence. Traditional approaches to Part-Of-Speech Tagging have relied on statistical models like Hidden Markov Models (HMMs) and Conditional Random Fields (CRFs). In contrast, modern techniques employ neural architectures such as BiLSTMs or Transformers, often pre-trained on large-scale corpora and fine-tuned for tagging tasks. These models can capture both local and global dependencies in a sentence, improving tagging accuracy, especially in morphologically rich languages. In multilingual and cross-lingual settings, Part-Of-Speech Tagging presents additional challenges due to typological variation, tokenization inconsistencies and the limited availability of annotated resources for low-resource languages. Prompt-based approaches using LLMs have recently emerged as a promising alternative, particularly for zero-shot and few-shot scenarios, where task-specific annotations are scarce or unavailable. In the context of our survey, we found two datasets which are XGLUE-POS \cite{liang2020xglue}, \href{https://lindat.mff.cuni.cz/repository/xmlui/handle/11234/1-2837}{UDPOS} and XTREME \cite{hu2020xtreme}. Table \ref{tab:pos} compiles these relevant datasets, the prompting strategies explored, the number of languages evaluated, associated research efforts and the best-performing method identified across multilingual settings.

\footnotesize
\begin{longtable}{p{.6in}p{.9in}p{1.4in}p{.5in}p{.8in}p{.8in}}
    \caption{Prompt Engineering Analysis for Code-Mixing Language Identification Task}
    \label{tab:cmlid} \\
    \toprule
    \textbf{Dataset} & \textbf{Prompting Strategy} & \textbf{LLM(s)}  & \textbf{SoTA} & \textbf{\#Language Pairs} & \textbf{References} \\
    \midrule
    \endfirsthead
    
    \multicolumn{6}{c}%
    {{\bfseries \tablename\ \thetable{} -- continued from previous page}} \\
    \toprule
    \textbf{Dataset} & \textbf{Prompting Strategy} & \textbf{LLM(s)}  & \textbf{SoTA} & \textbf{\#Language Pairs} & \textbf{References} \\
    \midrule
    \endhead

    \midrule \multicolumn{6}{c}{{Continued on next page}} \\
    \endfoot
    
    \bottomrule
    \endlastfoot

    LinCE & En-Basic, En-Basic + Variations   & GPT-3.5-Turbo, BLOOMZ-560M, BLOOMZ-1.1B, BLOOMZ-1.7B, BLOOMZ-3B, BLOOMZ-7.1B, mT0-300M, mT0-580M, mT0-1.2B, mT0-3.7B, mT0-13B, XGLM-564M, XGLM-1.7B, XGLM-2.9B, XGLM-4.5B, XGLM-7.5B     & En-Basic + Variations  & 2 & \cite{zhang2023multilingual}\\

\end{longtable}



\footnotesize
\begin{longtable}{p{.6in}p{.9in}p{1.4in}p{.5in}p{.8in}p{.8in}}
    \caption{Prompt Engineering Analysis for Part-Of-Speech Tagging Task}
    \label{tab:pos} \\
    \toprule
    \textbf{Dataset} & \textbf{Prompting Strategy} & \textbf{LLM(s)}  & \textbf{SoTA} & \textbf{\#Languages} & \textbf{References} \\
    \midrule
    \endfirsthead
    
    \multicolumn{6}{c}%
    {{ \tablename\ \thetable{} Continued from the Previous Page}} \\
    \toprule
    \textbf{Dataset} & \textbf{Prompting Strategy} & \textbf{LLM(s)}  & \textbf{SoTA} & \textbf{\#Languages} & \textbf{References} \\
    \midrule
    \endhead

    \multicolumn{6}{c}{{\tablename\ \thetable{} Continued on the Next Page}} \\
    \endfoot
    
    \bottomrule
    \endlastfoot

    XGLUE-POS & En-Basic    & GPT-3.5-Turbo    & En-Basic & 17 & \cite{lai2023chatgpt} \\
        \midrule
        XTREME & Iterative Prompting, Decomposed Prompting (Generative), Decomposed Prompting (Probabilistic)     & BLOOMZ-7.1B, mTk-Instruct-13B, Mistral-7B, LLaMA-2-13B, LLaMA-2-7B     & Decomposed Prompting (Probabilistic) & 38 & \cite{nie2024decomposed} \\
    \midrule
        UDPOS & Native-Basic     & GPT-3.5-Turbo, GPT-4, GPT-3.5 (Text-Davinci-003), XGLM, BLOOMZ & Native-Basic & 38 & \cite{ahuja2023mega} \\

\end{longtable}

\normalsize
\subsection{Named Entity Recognition}

Named Entity Recognition is an information extraction task in Natural NLP that focuses on identifying and classifying named entities such as persons, organizations, locations, dates and other predefined categories within unstructured text. Traditionally, Named Entity Recognition has been addressed using statistical models like Conditional Random Fields (CRFs) and sequence-based neural architectures such as BiLSTMs. With the advent of pre-trained transformer models like BERT and mBERT, Named Entity Recognition systems have significantly improved, especially in capturing contextual cues around entities. Multilingual Named Entity Recognition is particularly challenging due to differences in orthography, morphology, word order and entity naming conventions across languages. Additionally, low-resource languages often suffer from a lack of annotated corpora. Prompt-based learning using LLMs has recently gained traction for tackling Named Entity Recognition in zero-shot and few-shot settings. These methods reformulate the Named Entity Recognition task as a question-answering or span-extraction problem, allowing LLMs to detect entities without requiring extensive task-specific training. Our survey identified several datasets including MultiCoNER \cite{malmasi2022semeval}, CoNLLPP \cite{wang2019crossweigh}, MSRA \cite{levow2006third}, WikiAnn \cite{pan2017cross} and MasakhaNER \cite{adelani2021masakhaner} and prompting strategies that explore multilingual Named Entity Recognition. Table~\ref{tab:ner} lists these datasets, the range of languages they encompass, the prompting techniques applied, key studies in the domain and the best-performing approaches observed across various multilingual settings.

\footnotesize
\begin{longtable}{p{.6in}p{.9in}p{1.4in}p{.5in}p{.8in}p{.8in}}
    \caption{Prompt Engineering Analysis for Named Entity Recognition Task}
    \label{tab:ner} \\
    \toprule
    \textbf{Dataset} & \textbf{Prompting Strategy} & \textbf{LLM(s)}  & \textbf{SoTA} & \textbf{\#Languages} & \textbf{References} \\
    \midrule
    \endfirsthead
    
    \multicolumn{6}{c}%
    {{ \tablename\ \thetable{} Continued from the Previous Page}} \\
    \toprule
    \textbf{Dataset} & \textbf{Prompting Strategy} & \textbf{LLM(s)}  & \textbf{SoTA} & \textbf{\#Languages} & \textbf{References} \\
    \midrule
    \endhead

    \multicolumn{6}{c}{{\tablename\ \thetable{} Continued on the Next Page}} \\ \\
    \endfoot
    
    \bottomrule
    \endlastfoot

    MultiCoNER & En-Basic    & GPT-3.5-Turbo    & En-Basic & 5 & \cite{adelani2021masakhaner} \\
        \midrule
        CoNLLPP & En-Basic, ChatIE     & GPT-3.5-Turbo, InstructGPT (Text-Davinci-003), LLaMA-2-7B-Chat, ChatGLM2-6B     & ChatIE & 1 & \cite{wei2023chatie} \\
        \midrule
        MSRA & En-Basic, ChatIE     & GPT-3.5-Turbo, InstructGPT (Text-Davinci-003), LLaMA-2-7B-Chat, ChatGLM2-6B     & ChatIE & 1 & \cite{wei2023chatie} \\
        \midrule
        WikiAnn & X-Basic, Native-Basic, En-Basic    & BLOOMZ-7.1B, BLOOM-7.1B, GPT-3.5-Turbo, mT5-13B, mT0-13B, GPT-4, GPT-3.5 (Text-Davinci-003), XGLM, BLOOMZ     & X-Basic & 56 & \cite{asai2023buffet, ahuja2023mega} \\
        \midrule
        MasakhaNER & X-Basic, Native-Basic, En-Basic     & BLOOMZ-7.1B, BLOOM-7.1B, GPT-3.5-Turbo, mT5-13B, mT0-13B     & X-Basic, Native-Basic  & 9 & \cite{asai2023buffet} \\

\end{longtable}

\normalsize
\subsection{Event Extraction}

Event Extraction is an NLP task focused on identifying and classifying events described in a text, as well as extracting the arguments associated with those events, such as participants, time, location and other relevant information. Deep learning models, particularly transformer-based architectures like BERT and RoBERTa, have been shown to improve performance on Event Extraction by leveraging large-scale pre-training on diverse datasets. Event Extraction systems are usually trained on annotated corpora where events and their arguments are manually labeled. However, these systems often face challenges in handling ambiguous or underspecified events, different event types and fine-grained argument roles, which can vary significantly across languages and domains. Multilingual Event Extraction introduces added complexity due to differences in syntactic structures, event semantics and vocabulary across languages. For example, certain event types or argument roles might not have direct counterparts in all languages, making cross-lingual transfer a non-trivial challenge. Models like mBERT, XLM-R, and others that utilize cross-lingual representations are key to tackling these challenges, although the quality of event extraction is often influenced by the availability of annotated data in the target languages. In our survey, we identified two relevant datasets DuEE \cite{li2020duee} and ACE05 \cite{walker2006ace}. Table \ref{tab:eventextract}  mentions these datasets, detailing the different prompting methods explored, the number of languages covered, the relevant studies and the SoTA prompting technique.

\footnotesize
\begin{longtable}{p{.6in}p{.9in}p{1.4in}p{.5in}p{.8in}p{.8in}}
    \caption{Prompt Engineering Analysis for Event Extraction Task}
    \label{tab:eventextract} \\
    \toprule
    \textbf{Dataset} & \textbf{Prompting Strategy} & \textbf{LLM(s)}  & \textbf{SoTA} & \textbf{\#Languages} & \textbf{References} \\
    \midrule
    \endfirsthead
    
    \multicolumn{6}{c}%
    {{\bfseries \tablename\ \thetable{} -- continued from previous page}} \\
    \toprule
    \textbf{Dataset} & \textbf{Prompting Strategy} & \textbf{LLM(s)}  & \textbf{SoTA} & \textbf{\#Languages} & \textbf{References} \\
    \midrule
    \endhead

    \midrule \multicolumn{6}{c}{{Continued on next page}} \\
    \endfoot
    
    \bottomrule
    \endlastfoot

    DuEE & En-Basic, ChatIE    & GPT-3.5-Turbo, InstructGPT (Text-Davinci-003), LLaMA-2-7B-Chat, ChatGLM2-6B     & ChatIE & 1 & \cite{wei2023chatie} \\
    \midrule
    ACE05 & En-Basic, ChatIE     & GPT-3.5-Turbo, InstructGPT (Text-Davinci-003), LLaMA-2-7B-Chat, ChatGLM2-6B     & ChatIE & 1 & \cite{wei2023chatie} \\

\end{longtable}

\normalsize
\subsection{Word Sense Disambiguation}

Word Sense Disambiguation in NLP involves determining the correct meaning (or sense) of a word within a specific context, especially when the word has multiple possible meanings. Traditionally, Word Sense Disambiguation has been approached using supervised learning with sense-annotated corpora, unsupervised methods leveraging sense inventories like WordNet, or knowledge-based techniques that make use of lexical resources. Recent advancements have seen a shift toward using contextual embeddings from transformer-based models, such as BERT and XLM-R, to model sense distinctions. In multilingual and cross-lingual settings, Word Sense Disambiguation becomes even more complex due to differences in lexical sense granularity, alignment of sense inventories across languages and limited availability of annotated data in low-resource languages. Prompt-based methods with LLMs have introduced a promising zero-shot and few-shot paradigm for Word Sense Disambiguation. These techniques often reformulate Word Sense Disambiguation as a multiple-choice or span prediction problem, enabling models to leverage their internal knowledge without requiring extensive fine-tuning. We encountered only one dataset XL-WiC \cite{raganato2020xl} in our journey to prepare this survey. Table~\ref{tab:wsd} provides an overview of this dataset, the number of languages included, the prompting approaches examined, associated literature and the highest-performing strategy.

\normalsize
\subsection{Co-Reference Resolution}

Co-Reference Resolution in NLP involves identifying and linking expressions within a text that refer to the same entity. This typically includes resolving pronouns, noun phrases and other referring expressions to their corresponding entities. The Co-Reference Resolution task typically involves both syntactic and semantic analysis, as resolving references often requires an understanding of the context in which the reference is made. Many approaches to Co-Reference Resolution rely on supervised machine learning models, such as neural networks, which are trained on large annotated datasets where mentions and their co-references are labeled. Additionally, recent methods have integrated transformer-based models, such as BERT, which capture contextual relationships and can resolve references in a more flexible and context-aware manner. In multilingual and cross-lingual contexts, Co-Reference Resolution poses additional challenges due to linguistic variations in how co-reference is expressed. Some languages may use different grammatical structures, such as gendered pronouns or may omit explicit subjects or objects, which adds complexity to the resolution task. Despite these challenges, recent research has shown promising results by leveraging multilingual pre-trained models (e.g., mBERT or XLM-R) and cross-lingual transfer learning, allowing systems to generalize across languages with fewer language-specific resources. In our survey, we identified one key dataset relevant to Co-Reference Resolution, including XWinograd \cite{muennighoff2022crosslingual}. Table \ref{tab:coref} mentions this dataset, the different prompting methods explored, the number of languages covered and the SoTA prompting approach.

\footnotesize
\begin{longtable}{p{.6in}p{.9in}p{1.4in}p{.5in}p{.8in}p{.8in}}
    \caption{Prompt Engineering Analysis for Word Sense Disambiguation Task}
    \label{tab:wsd} \\
    \toprule
    \textbf{Dataset} & \textbf{Prompting Strategy} & \textbf{LLM(s)}  & \textbf{SoTA} & \textbf{\#Languages} & \textbf{References} \\
    \midrule
    \endfirsthead
    
    \multicolumn{6}{c}%
    {{\bfseries \tablename\ \thetable{} -- continued from previous page}} \\
    \toprule
    \textbf{Dataset} & \textbf{Prompting Strategy} & \textbf{LLM(s)}  & \textbf{SoTA} & \textbf{\#Languages} & \textbf{References} \\
    \midrule
    \endhead

    \midrule \multicolumn{6}{c}{{Continued on next page}} \\
    \endfoot
    
    \bottomrule
    \endlastfoot

    XL-WiC & Native-Basic,  En-CoT     & Codex (Code-Davinci-002), PaLM-540B     & Native-Basic & 12 & \cite{shi2022language} \\

\end{longtable}

\footnotesize
\begin{longtable}{p{.6in}p{.9in}p{1.4in}p{.5in}p{.8in}p{.8in}}
    \caption{Prompt Engineering Analysis for Coreference Resolution Task}
    \label{tab:coref} \\
    \toprule
    \textbf{Dataset} & \textbf{Prompting Strategy} & \textbf{LLM(s)}  & \textbf{SoTA} & \textbf{\#Languages} & \textbf{References} \\
    \midrule
    \endfirsthead
    
    \multicolumn{6}{c}%
    {{\bfseries \tablename\ \thetable{} -- continued from previous page}} \\
    \toprule
    \textbf{Dataset} & \textbf{Prompting Strategy} & \textbf{LLM(s)}  & \textbf{SoTA} & \textbf{\#Languages} & \textbf{References} \\
    \midrule
    \endhead

    \midrule \multicolumn{6}{c}{{Continued on next page}} \\
    \endfoot
    
    \bottomrule
    \endlastfoot

    XWinograd & X-Basic, Native-Basic, En-Basic     & BLOOMZ-7.1B, BLOOM-7.1B, GPT-3.5-Turbo, mT5-13B, mT0-13B     & Native-Basic & 4 & \cite{asai2023buffet}\\

\end{longtable}

\normalsize
\subsection{Relation Extraction}

Relation Extraction is a crucial task in NLP that involves identifying and classifying semantic relationships between entities within a text. Given a sentence or a document, the objective is to extract pairs of entities and the relations that hold between them. Traditional methods relied heavily on supervised learning and hand-labeled datasets but the emergence of deep learning and transformer-based models has revolutionized the task, leading to significant improvements in performance. Models like BERT and its variants have enabled fine-tuning for specific Relation Extraction tasks with minimal task-specific architecture modifications. In multilingual settings, Relation Extraction presents additional challenges, such as language-specific syntactic and semantic variations, as well as handling linguistic structures that may not directly correspond to the relation types in other languages. Despite these challenges, multilingual models like mBERT and XLM-R have shown promise in transferring relational knowledge across languages, though the quality of Relation Extraction may degrade for low-resource languages or non-standardized relation formats. Our exploration identified three benchmarks which are NYT11-HRL \cite{takanobu2019hierarchical}, DuIE2.0 \cite{li2019duie} and SMiLER \cite{seganti2021multilingual}. Table \ref{tab:re} lists these datasets, the different prompting methods explored, the number of languages tested, associated research contributions and the best-performing prompting approaches.

\normalsize
\subsection{Natural Language Inference}

Natural Language Inference involves determining the inferential relationship between a pair of sentences: a premise and a hypothesis. The goal is to classify whether the hypothesis is entailed by the premise, contradicts it, or is neutral with respect to it. Natural Language Inference serves as a benchmark for testing deeper levels of language comprehension and reasoning in LLMs. Success on this task requires mastery over various linguistic phenomena, including paraphrasing, negation, co-reference, temporal ordering and world knowledge. Recent advances in LLMs and prompting methods have enabled zero-shot and few-shot approaches to Natural Language Inference, where models infer relationships between sentence pairs with little to no task-specific training. These approaches often rely on natural language prompts to elicit entailment behavior, with varied success depending on the model’s training corpus and alignment quality. In multilingual contexts, Natural Language Inference becomes significantly more complex. Challenges include capturing nuanced relationships expressed differently across languages, aligning semantic equivalences despite syntactic divergences and transferring inference capabilities to low-resource languages. Cross-lingual transfer techniques and multilingual pretraining have enabled improvements in this area but robust and culturally aware inference remains an open problem. Our exploration identified several benchmarks including XNLI \cite{conneau2018xnli}, AmericasNLI \cite{ebrahimi2021americasnli}, PARSINLU \cite{khashabi2021parsinlu}, OCNLI \cite{hu2020ocnli}, KLUE \cite{park2105klue} and IndicXNLI\cite{aggarwal2022indicxnli}. Table \ref{tab:nli} showcases the datasets reviewed, the prompting paradigms adopted, language coverage, related publications and the most effective prompting configurations observed in the literature.

\footnotesize
\begin{longtable}{p{.6in}p{.9in}p{1.4in}p{.5in}p{.8in}p{.8in}}
    \caption{Prompt Engineering Analysis for Relation Extraction Task}
    \label{tab:re} \\
    \toprule
    \textbf{Dataset} & \textbf{Prompting Strategy} & \textbf{LLM(s)}  & \textbf{SoTA} & \textbf{\#Languages} & \textbf{References} \\
    \midrule
    \endfirsthead
    
    \multicolumn{6}{c}%
    {{\bfseries \tablename\ \thetable{} -- continued from previous page}} \\
    \toprule
    \textbf{Dataset} & \textbf{Prompting Strategy} & \textbf{LLM(s)}  & \textbf{SoTA} & \textbf{\#Languages} & \textbf{References} \\
    \midrule
    \endhead

    \midrule \multicolumn{6}{c}{{Continued on next page}} \\
    \endfoot
    
    \bottomrule
    \endlastfoot

    NYT11-HRL & En-Basic, ChatIE    & GPT-3.5-Turbo, InstructGPT (Text-Davinci-003), LLaMA-2-7B-Chat, ChatGLM2-6B     & ChatIE & 1 & \cite{wei2023chatie} \\
    \midrule
    DuIE2.0 & En-Basic, ChatIE     & GPT-3.5-Turbo, InstructGPT (Text-Davinci-003), LLaMA-2-7B-Chat, ChatGLM2-6B     & ChatIE & 1 & \cite{wei2023chatie} \\
    \midrule
    SMiLER & En-Basic    & GPT-3.5-Turbo     & En-Basic & 14 & \cite{lai2023chatgpt} \\

\end{longtable}

\normalsize
\subsection{Code-Mixing Natural Language Inference}

Code-Mixing Natural Language Inference in NLP refers to determining the inferential relationship between a pair of code-mixed sentences, typically drawn from linguistically blended utterances where two or more languages are intermixed within or across the premise and hypothesis. The task aims to classify whether the hypothesis is entailed by the premise, contradicts it or is neutral with respect to it. Code-Mixing Natural Language Inference serves as a benchmark for evaluating language understanding, cross-lingual reasoning and cultural context interpretation in multilingual communities where code-mixing is a natural mode of communication. Unlike standard Natural Language Inference, Code-Mixing Natural Language Inference presents additional complexity due to lexical, syntactic and semantic interference between the embedded languages. It requires not only traditional reasoning capabilities such as paraphrasing, negation or world knowledge, but also an ability to resolve expressions across language switch points. Code-mixing poses challenges for traditional monolingual Natural Language Inference models, which often fail to interpret or align meaning across languages. Recent research has explored leveraging multilingual pretrained transformers, cross-lingual embeddings and data augmentation techniques to improve performance in code-mixed scenarios. Prompt-based few-shot or zero-shot learning has also shown promise, especially when prompts are crafted to reflect the sociolinguistic norms of the mixed languages involved. Our survey includes one identified dataset GLUECoS-NLI \cite{khanuja2020gluecos}. Table \ref{tab:cmnli} presents a comparative overview of this dataset reviewed, prompting configurations explored, language combinations covered and evaluation findings across the literature.

\footnotesize
\begin{longtable}{p{.6in}p{.9in}p{1.4in}p{.5in}p{.8in}p{.8in}}
    \caption{Prompt Engineering Analysis for Natural Language Inference Task}
    \label{tab:nli} \\
    \toprule
    \textbf{Dataset} & \textbf{Prompting Strategy} & \textbf{LLM(s)}  & \textbf{SoTA} & \textbf{\#Languages} & \textbf{References} \\
    \midrule
    \endfirsthead

    \multicolumn{6}{c}%
    {{ \tablename\ \thetable{} Continued from the Previous Page}} \\
    
    \toprule
    \textbf{Dataset} & \textbf{Prompting Strategy} & \textbf{LLM(s)}  & \textbf{SoTA} & \textbf{\#Languages} & \textbf{References} \\
    \midrule
    \endhead

    \multicolumn{6}{c}{{ \tablename\ \thetable{} Continued on the Next Page}}
    
    \endfoot
    
    \bottomrule
    \endlastfoot

        AmericasNLI & X-Basic, Native-Basic, En-Basic     & BLOOMZ-7.1B, BLOOM-7.1B, GPT-3.5-Turbo, mT5-13B, mT0-13B     & X-Basic & 10 & \cite{asai2023buffet} \\
        \midrule
        PARSINLU & X-Basic, Native-Basic, En-Basic     & BLOOMZ-7.1B, BLOOM-7.1B, GPT-3.5-Turbo, mT5-13B, mT0-13B     & X-Basic & 1 & \cite{asai2023buffet} \\
        \midrule
        OCNLI & X-Basic, Native-Basic, En-Basic     & BLOOMZ-7.1B, BLOOM-7.1B, GPT-3.5-Turbo, mT5-13B, mT0-13B     & En-Basic & 1 & \cite{asai2023buffet} \\
        \midrule
        KLUE & X-Basic, Native-Basic, En-Basic     & BLOOMZ-7.1B, BLOOM-7.1B, GPT-3.5-Turbo, mT5-13B, mT0-13B     & X-Basic & 1 & \cite{asai2023buffet} \\
        \midrule
        IndicXNLI & Native-Basic, X-Basic, Translate-En    & GPT-3.5-Turbo, GPT-4, GPT-3.5 (Text-Davinci-003), XGLM, BLOOMZ    & Translate-En & 11 & \cite{ahuja2023mega} \\
        \midrule
        
        XNLI & Native-CoT, Translate-En-CoT, En-Basic, En-CoT, Translate-En, XLT, CLP, X-Basic, Native-Basic    & BLOOM-0.6B, BLOOM-1.1B, BLOOM-1.7B, BLOOM-3B, BLOOM-7.1B, BLOOMZ-7.1B, Codex (Code-Davinci-002), GPT-3 (Text-Davinci-002), GPT-3.5 ( Text-Davinci-003), GPT-3.5-Turbo, GPT-4, LLaMA-13B, LLaMA-2-13B, LLaMA-2-13B-Chat, LLaMA-2-7B, LLaMA-2-70B-Chat, LLaMA-30B, LLaMA-7B, Mistral-7B-Instruct, OpenLLaMA-13B, OpenLLaMA-3B, OpenLLaMA-7B, OpenLLaMA-V2-3B, OpenLLaMA-V2-7B, PaLM-540B, PolyLM-13B, PolyLM-1B, RedPajama-3B, RedPajama-7B, XGLM-1.7B, XGLM-2.9B, XGLM-564M, XGLM-7.5B, mT0-13B, mT5-13B, BLOOMZ, XGLM     & XLT, Translate-En  & 15 & \cite{etxaniz2023multilingual, huang2023not, qin2023cross, kim2023cross, liu2024translation, lai2023chatgpt, asai2023buffet, ahuja2023mega} \\

\end{longtable}

\normalsize
\subsection{Dialogue Evaluation}

Dialogue Evaluation is an essential task in NLP that focuses on assessing the quality of responses generated by conversational agents. Unlike traditional classification or generation tasks, dialogue evaluation requires nuanced judgments across multiple dimensions, such as coherence, relevance, fluency, informativeness and appropriateness in context. The complexity of this task arises from the open-ended nature of dialogue, the subjectivity of human evaluation and the challenge of capturing conversational dynamics over multiple turns. Multilingual dialogue evaluation introduces further challenges, such as ensuring consistency across languages, adapting evaluation frameworks to different conversational norms and addressing the scarcity of high-quality annotated datasets in low-resource languages. Addressing these challenges is critical for building inclusive and globally effective conversational Artificial Intelligence (AI) systems. For this task, several multilingual datasets have been explored in the literature, including FED \cite{mehri2020unsupervised}, SEE \cite{see2019makes}, KdConv \cite{zhou2020kdconv}, LCCC \cite{wang2020large} and DSTC-11 \cite{rodriguez2023overview}. These datasets, which offer diverse scenarios for dialogue generation and evaluation, provide a range of multilingual resources for assessing dialogue quality. Table \ref{tab:dias} details these datasets, the prompting methods applied, the number of languages experimented with and the most effective (SoTA) prompting strategies identified for multilingual dialogue evaluation.

\footnotesize
\begin{longtable}{p{.6in}p{.9in}p{1.4in}p{.5in}p{.8in}p{.8in}}
    \caption{Prompt Engineering Analysis for Code-Mixing Natural Language Inference Task}
    \label{tab:cmnli} \\
    \toprule
    \textbf{Dataset} & \textbf{Prompting Strategy} & \textbf{LLM(s)}  & \textbf{SoTA} & \textbf{\#Language Pairs} & \textbf{References} \\
    \midrule
    \endfirsthead

    \multicolumn{6}{c}%
    {{ \tablename\ \thetable{} Continued from the Previous Page}} \\
    
    \toprule
    \textbf{Dataset} & \textbf{Prompting Strategy} & \textbf{LLM(s)}  & \textbf{SoTA} & \textbf{\#Language Pairs} & \textbf{References} \\
    \midrule
    \endhead

    \multicolumn{6}{c}{{ \tablename\ \thetable{} Continued on the Next Page}} 
    \endfoot
    
    \bottomrule
    \endlastfoot

        GLUECoS-NLI & Native-Basic    & GPT-3.5-Turbo, GPT-4, GPT-3.5 (Text-Davinci-003), XGLM, BLOOMZ    & Native-Basic & 1 & \cite{ahuja2023mega} \\

\end{longtable}

\footnotesize
\begin{longtable}{p{.6in}p{.9in}p{1.4in}p{.5in}p{.8in}p{.8in}}
    \caption{Prompt Engineering Analysis for Dialogue Evaluation Task}
    \label{tab:dias} \\
    \toprule
    \textbf{Dataset} & \textbf{Prompting Strategy} & \textbf{LLM(s)}  & \textbf{SoTA} & \textbf{\#Languages} & \textbf{References} \\
    \midrule
    \endfirsthead
    
    \multicolumn{6}{c}%
    {{ \tablename\ \thetable{} Continued from the Previous Page}} \\
    \toprule
    \textbf{Dataset} & \textbf{Prompting Strategy} & \textbf{LLM(s)}  & \textbf{SoTA} & \textbf{\#Languages} & \textbf{References} \\
    \midrule
    \endhead

    \multicolumn{6}{c}{{ \tablename\ \thetable{} Continued on the Next Page}} 
    \endfoot
    
    \bottomrule
    \endlastfoot

    FED  & En-Basic, HD Prompting, MEEP, MEEP + Variations, G-Eval, GPTScore   & GPT-3.5 (Text-Davinci-003), GPT-3.5-Turbo, LLaMA-7B      & MEEP + Variations & 3 & \cite{ferron2023meep} \\
    \midrule
    SEE  & En-Basic, HD Prompting, MEEP, MEEP + Variations, G-Eval, GPTScore    & GPT-3.5 (Text-Davinci-003), GPT-3.5-Turbo, LLaMA-7B      & MEEP + Variations & 3 & \cite{ferron2023meep} \\
    \midrule
    KdConv  & En-Basic, HD Prompting, MEEP, MEEP + Variations, G-Eval, GPTScore   & GPT-3.5 (Text-Davinci-003), GPT-3.5-Turbo, LLaMA-7B      & MEEP + Variations & 2 & \cite{ferron2023meep} \\
    \midrule
    DSTC-11  & En-Basic    & GPT-3.5-Turbo     & En-Basic & 3 & \cite{mendoncca2023simple} \\
    \midrule
    LCCC  & En-Basic, HD Prompting, MEEP, MEEP + Variations, G-Eval, GPTScore    & GPT-3.5 (Text-Davinci-003), GPT-3.5-Turbo, LLaMA-7B      & MEEP + Variations & 2 & \cite{ferron2023meep} \\

\end{longtable}

\normalsize
\subsection{Task Understanding Consistency}

Task Understanding Consistency in NLP refers to the ability of a model to consistently and accurately comprehend the requirements of different tasks across various contexts, ensuring reliable performance despite changes in task formulations, linguistic nuances or domain variations. This task evaluates how well a model maintains the understanding of task-specific instructions and can adapt its behavior accordingly across different conditions or languages, without being influenced by irrelevant or ambiguous inputs. In multilingual settings, Task Understanding Consistency becomes more critical due to the variations in how tasks are expressed across different languages. Different languages may have unique sentence structures, word orders or syntactic constructions that could confuse a model if it is not trained to handle these differences. Furthermore, language-specific cultural or contextual nuances can influence task interpretations, posing additional challenges for models attempting to perform tasks consistently across languages. In the context of multilingual NLP, research has explored various methods to improve task understanding, such as the use of multilingual pre-trained models like mBERT and XLM-R, which aim to provide a shared understanding of tasks across languages. Another approach involves incorporating explicit task representations or instructions into the training process, helping models recognize task-specific features more effectively. In our work, we found two relevant datasets XNLI \cite{conneau2018xnli} and PAWS-X\cite{yang2019paws}.  Table \ref{tab:mtuc} lists the datasets found, the number of languages involved, the prompting methods used, relevant studies and the best-performing prompting technique.

\footnotesize
\begin{longtable}{p{.6in}p{.9in}p{1.4in}p{.5in}p{.8in}p{.8in}}
    \caption{Prompt Engineering Analysis for Task Understanding Consistency Task}
    \label{tab:mtuc} \\
    \toprule
    \textbf{Dataset} & \textbf{Prompting Strategy} & \textbf{LLM(s)}  & \textbf{SoTA} & \textbf{\#Language Pairs} & \textbf{References} \\
    \midrule
    \endfirsthead
    
    \multicolumn{6}{c}%
    {{\bfseries \tablename\ \thetable{} -- continued from previous page}} \\
    \toprule
    \textbf{Dataset} & \textbf{Prompting Strategy} & \textbf{LLM(s)}  & \textbf{SoTA} & \textbf{\#Language Pairs} & \textbf{References} \\
    \midrule
    \endhead

    \midrule \multicolumn{6}{c}{{Continued on next page}} \\
    \endfoot
    
    \bottomrule
    \endlastfoot

    XNLI & En-Basic    & GPT-3.5-Turbo    & En-Basic & 4 & \cite{ohmer2023evaluating}\\
    \midrule

    PAWS-X & En-Basic    & GPT-3.5-Turbo    & En-Basic & 4 & \cite{ohmer2023evaluating}\\

\end{longtable}



    

    



\normalsize
\subsection{Summarization}

Summarization is a key NLP task that focuses on generating concise and coherent summaries that retain the essential information from longer input texts. The task requires not only surface-level understanding but also deeper semantic comprehension to determine which parts of the text are most informative. Recent transformer-based language models like BART, T5 and mT5 have significantly advanced the performance of summarization systems, enabling both zero-shot and few-shot capabilities across different domains and languages. Multilingual Summarization introduces further complexities due to linguistic diversity, such as syntactic variation, differences in discourse structure and cultural specificity in what is considered salient. Pretrained multilingual models and instruction-tuned LLMs have been increasingly applied to address these challenges, often showing strong performance even in unseen languages. In the course of our survey, the datasets we found includes XL-Sum \cite{hasan2021xl} and CLCTS \cite{zhang2023cross}. Table \ref{tab:summa} highlights these datasets, the prompting techniques utilized, number of languages evaluated across different experiments, pertinent research and the prompting configurations that yielded the best performance.

\footnotesize
\begin{longtable}{p{.6in}p{.9in}p{1.4in}p{.5in}p{.8in}p{.8in}}
    \caption{Prompt Engineering Analysis for Summarization Task}
    \label{tab:summa} \\
    \toprule
    \textbf{Dataset} & \textbf{Prompting Strategy} & \textbf{LLM(s)}  & \textbf{SoTA} & \textbf{\#Languages} & \textbf{References} \\
    \midrule
    \endfirsthead
    
    \multicolumn{6}{c}%
    {{ \tablename\ \thetable{} Continued from the Previous Page}} \\
    \toprule
    \textbf{Dataset} & \textbf{Prompting Strategy} & \textbf{LLM(s)}  & \textbf{SoTA} & \textbf{\#Languages} & \textbf{References} \\
    \midrule
    \endhead

    \multicolumn{6}{c}{{ \tablename\ \thetable{} Continued on the Next Page}}
    \endfoot
    
    \bottomrule
    \endlastfoot

    XL-Sum & En-Basic, X-Basic, En-CoT, XLT, Native-Basic, Native-CoT, Translate-En-CoT, Translate-En       & GPT-3.5 ( Text-Davinci-003), GPT-3.5-Turbo, GPT-4, XGLM, BLOOMZ LLaMA-2-70B-Chat, Codex (Code-Davinci-002), Mistral-7B-Instruct, LLaMA-2-13B-Chat, BLOOMZ-7.1B, BLOOM-7.1B, mT5-13B, mT0-13B, PaLM 2-S,PaLM 2-L      & En-Basic, Translate-En-CoT & 47 & \cite{huang2023not, liu2024translation, lai2023chatgpt, asai2023buffet, intrator2024breaking, ahuja2023mega} \\
    \midrule

    CLCTS & En-Basic, En-Basic + Variations, Native-Basic, Native-Basic + Variations     & GPT-3.5-Turbo      & En-Basic, Native-Basic & 2 & \cite{zhang2023cross} \\

\end{longtable}

\normalsize
\subsection{Code-Mixing Summarization}

Code-Mixing Summarization aims to generate concise and coherent summaries from code-mixed input texts, texts that contain lexical and syntactic elements from two or more languages within the same utterance or document. Unlike traditional summarization tasks conducted in monolingual settings, this task requires models to effectively handle intra-sentential language switching, transliteration, orthographic variation, and non-standard grammar structures characteristic of informal bilingual or multilingual communication. This task poses unique challenges for LLMs due to the scarcity of annotated code-mixed summarization datasets, the lack of syntactic regularity in inputs and limited cross-lingual generalization capabilities in low-resource languages. Our survey identified one notable dataset Gupshup \cite{mehnaz2021gupshup} associated with this task. Table \ref{tab:cmsummary} lists this dataset along with the prompting methods explored, the number of language combinations involved, references to relevant studies and the prompting strategy that demonstrated the strongest performance in this domain.

\normalsize
\subsection{Paraphrasing}

Paraphrasing is a task centered on generating alternative expressions of a given sentence or phrase while preserving its original meaning. Paraphrasing requires a nuanced understanding of lexical, syntactic and semantic variations to ensure that the generated output remains semantically faithful to the input yet introduces sufficient linguistic diversity. This task can be framed in multiple ways such as one-to-one sentence rewriting, generating multiple paraphrases, or controlled paraphrasing where specific constraints (e.g., style, formality or length) are imposed. Traditional approaches to paraphrasing relied on rule-based methods or statistical machine translation techniques. However, the advent of pretrained transformer-based models 

\footnotesize
\begin{longtable}{p{.6in}p{.9in}p{1.4in}p{.5in}p{.8in}p{.8in}}
    \caption{Prompt Engineering Analysis for Code-Mixing Summarization Task}
    \label{tab:cmsummary} \\
    \toprule
    \textbf{Dataset} & \textbf{Prompting Strategy} & \textbf{LLM(s)}  & \textbf{SoTA} & \textbf{\#Language Pairs} & \textbf{References} \\
    \midrule
    \endfirsthead
    
    \multicolumn{6}{c}%
    {{\bfseries \tablename\ \thetable{} -- continued from previous page}} \\
    \toprule
    \textbf{Dataset} & \textbf{Prompting Strategy} & \textbf{LLM(s)}  & \textbf{SoTA} & \textbf{\#Language Pairs} & \textbf{References} \\
    \midrule
    \endhead

    \midrule \multicolumn{6}{c}{{Continued on next page}} \\
    \endfoot
    
    \bottomrule
    \endlastfoot

    Gupshup & En-Basic, En-Basic + Variations     & GPT-3.5-Turbo, BLOOMZ-560M, BLOOMZ-1.1B, BLOOMZ-1.7B, BLOOMZ-3B, BLOOMZ-7.1B, mT0-300M, mT0-580M, mT0-1.2B, mT0-3.7B, mT0-13B, XGLM-564M, XGLM-1.7B, XGLM-2.9B, XGLM-4.5B, XGLM-7.5B     & En-Basic + Variations & 1 & \cite{zhang2023multilingual}\\

\end{longtable}

\footnotesize
\begin{longtable}{p{.6in}p{.9in}p{1.4in}p{.5in}p{.8in}p{.8in}}
    \caption{Prompt Engineering Analysis for Paraphrasing Task}
    \label{tab:parap} \\
    \toprule
    \textbf{Dataset} & \textbf{Prompting Strategy} & \textbf{LLM(s)}  & \textbf{SoTA} & \textbf{\#Languages} & \textbf{References} \\
    \midrule
    \endfirsthead
    
    \multicolumn{6}{c}%
    {{\bfseries \tablename\ \thetable{} -- continued from previous page}} \\
    \toprule
    \textbf{Dataset} & \textbf{Prompting Strategy} & \textbf{LLM(s)}  & \textbf{SoTA} & \textbf{\#Languages} & \textbf{References} \\
    \midrule
    \endhead

    \midrule \multicolumn{6}{c}{{Continued on next page}} \\
    \endfoot
    
    \bottomrule
    \endlastfoot

    PAWS-X & Native-CoT, Translate-En-CoT, En-Basic, En-CoT, Translate-En, XLT, CLP, Native-Basic, X-Basic     & BLOOM-0.6B, BLOOM-1.7B, BLOOM-3B, BLOOM-7.1B, BLOOMZ-7.1B, Codex (Code-Davinci-002), GPT-3 (Text-Davinci-002), GPT-3.5 ( Text-Davinci-003), GPT-3.5-Turbo, GPT-4, LLaMA-13B, LLaMA-2-13B, LLaMA-2-13B-Chat, LLaMA-2-7B, LLaMA-2-70B-Chat, LLaMA-30B, LLaMA-7B, Mistral-7B-Instruct, OpenLLaMA-13B, OpenLLaMA-3B, OpenLLaMA-7B, OpenLLaMA-V2-3B, OpenLLaMA-V2-7B, PaLM-540B, PolyLM-13B, PolyLM-1B, RedPajama-3B, RedPajama-7B, XGLM-1.7B, XGLM-2.9B, XGLM-564M, XGLM-7.5B, mT0-13B, mT5-13B, XGLM, BLOOMZ     & XLT & 7 & \cite{etxaniz2023multilingual, huang2023not, qin2023cross, liu2024translation, asai2023buffet, ahuja2023mega} \\

\end{longtable}

\normalsize
(e.g., T5, BART, and GPT variants) has significantly enhanced the quality and fluency of paraphrased outputs through both supervised and prompt-based generation strategies. Paraphrasing in a multilingual context poses additional challenges due to the syntactic and idiomatic differences among languages. It demands not only the ability to restate information within a single language but also the adaptability to produce paraphrases in different languages or even across languages. Furthermore, generating natural paraphrases in low-resource languages necessitates models capable of effective cross-lingual transfer. Paraphrasing is a relatively under-explored task in multilingual prompting literature and we found only one relevant dataset PAWS-X \cite{yang2019paws}. Table \ref{tab:parap} lists the dataset, prompting techniques tested, language coverage, associated research efforts and the prompting strategy that achieved the best results.

\normalsize
\subsection{Social Bias}

Social Bias Detection is an important task in NLP aimed at identifying, analyzing and mitigating biased content in language models and datasets. This task involves detecting unfair, stereotypical or prejudiced associations in text that are based on social categories such as gender, race, ethnicity, religion, nationality, socioeconomic status or other demographic attributes. Social biases in NLP systems can propagate and amplify societal inequalities if not addressed, making this task essential for building responsible and ethical AI systems. The Social Bias task is generally framed as a classification or ranking problem where models are required to either detect the presence of bias in a given sentence or compare multiple outputs to determine which one is less biased. Some tasks may also require explanation generation or counterfactual reasoning to justify the prediction and improve transparency. Successful performance in this task requires models to understand nuanced social and cultural contexts and recognize implicit stereotypes, harmful generalizations or discriminatory phrasing. Prompt-based approaches have been used to evaluate and mitigate bias LLMs. For instance, bias can be tested by designing prompts that reveal stereotypical completions or differential sentiment associated with identity terms. Few-shot and zero-shot prompting have also enabled more interpretable bias evaluation in multilingual and code-mixed settings, where translation and language-switching further complicate the detection of social bias. In our work, we identified one dataset for this task which is WinoMT \cite{stanovsky2019evaluating}. Table \ref{tab:socialbias} summarizes the prompting methods explored for this task, datasets used, languages covered and research studies that contribute to our current understanding of social bias in multilingual LLMs.

\footnotesize
\begin{longtable}{p{.6in}p{.9in}p{1.4in}p{.5in}p{.8in}p{.8in}}
    \caption{Prompt Engineering Analysis for Social Bias Task}
    \label{tab:socialbias} \\
    \toprule
    \textbf{Dataset} & \textbf{Prompting Strategy} & \textbf{LLM(s)}  & \textbf{SoTA} & \textbf{\#Languages} & \textbf{References} \\
    \midrule
    \endfirsthead
    
    \multicolumn{6}{c}%
    {{\bfseries \tablename\ \thetable{} -- continued from previous page}} \\
    \toprule
    \textbf{Dataset} & \textbf{Prompting Strategy} & \textbf{LLM(s)}  & \textbf{SoTA} & \textbf{\#Languages} & \textbf{References} \\
    \midrule
    \endhead

    \midrule \multicolumn{6}{c}{{Continued on next page}} \\
    \endfoot
    
    \bottomrule
    \endlastfoot

    WinoMT & Native-Basic, Translate-En   & GPT-3.5-Turbo, GPT-4, GPT-3.5 (Text-Davinci-003), XGLM, BLOOMZ    & Native-Basic   & 8 & \cite{ahuja2023mega}\\

\end{longtable}

\normalsize
\subsection{Hope Detection}

Hope Detection is a specialized affective computing task in NLP that aims to identify expressions of hope in textual data. Unlike traditional sentiment analysis which typically categorizes text into broad emotional valence such as positive, negative or neutral, hope detection focuses on capturing a more nuanced psychological state associated with optimism, perseverance and positive expectations for the future. This task plays a significant role in mental health analysis, crisis management and understanding emotional resilience, particularly in social media contexts where individuals often express personal challenges, aspirations and coping mechanisms. Linguistically, detecting hope requires the model to go beyond surface-level sentiment cues and to interpret contextually grounded expressions that may involve implicit or culturally shaped notions of motivation, future-oriented thinking and emotional recovery. In multilingual and low-resource settings, the problem becomes even more complex due to variability in how hope is expressed across languages and cultures, as well as limited annotated data for training robust models. Additionally, the subjectivity of the emotion and overlaps with related affective states such as gratitude or determination necessitate fine-grained modeling approaches. Our survey highlights one dataset Hope \cite{garcia2024overview} central to this task. Table~\ref{tab:hopedetection} lists this dataset, the prompting techniques evaluated, the number of language settings covered, key studies conducted on this task and the prompting method that yielded the most promising performance.

\footnotesize
\begin{longtable}{p{.6in}p{.9in}p{1.4in}p{.5in}p{.8in}p{.8in}}
    \caption{Prompt Engineering Analysis for Hope Detection Task}
    \label{tab:hopedetection} \\
    \toprule
    \textbf{Dataset} & \textbf{Prompting Strategy} & \textbf{LLM(s)}  & \textbf{SoTA} & \textbf{\#Languages} & \textbf{References} \\
    \midrule
    \endfirsthead
    
    \multicolumn{6}{c}%
    {{\bfseries \tablename\ \thetable{} -- continued from previous page}} \\
    \toprule
    \textbf{Dataset} & \textbf{Prompting Strategy} & \textbf{LLM(s)}  & \textbf{SoTA} & \textbf{\#Languages} & \textbf{References} \\
    \midrule
    \endhead

    \midrule \multicolumn{6}{c}{{Continued on next page}} \\
    \endfoot
    
    \bottomrule
    \endlastfoot

    Hope & En-Basic, En-Basic + Variations, En-CoT    & GPT-3.5-Turbo    & En-Basic + Variations, En-CoT   & 2 & \cite{thuy2024empirical}\\

\end{longtable}

\normalsize
\subsection{Text Classification}

The Text Classification task focuses on assigning predefined labels or categories to text based on its content. This task can encompass several of the other tasks mentioned in this work, such as Named Entity Recognition, Hope Detection, etc. However, to keep the task definitions as distinct as possible for a clearer survey of prompting methods, we have included only those datasets under this task that could not be appropriately categorized under any of the other discussed tasks. The core challenge of Text Classification lies in understanding the content of the text and mapping it to one or more labels that best represent the conveyed information. While early methods focused on traditional machine learning algorithms such as Support Vector Machines (SVMs) and Naive Bayes, the advent of deep learning techniques, such as recurrent neural networks (RNNs) and transformers, has led to significant improvements in performance, particularly for more complex tasks. In the multilingual setting, Text Classification becomes even more challenging due to variations in syntax, semantics and cultural context across different languages. A model needs to be able to understand and classify text written in various languages without being explicitly trained for each language. Multilingual models, such as mBERT and XLM-R, have been designed to address these challenges by leveraging cross-lingual representations and shared knowledge, enabling a single model to perform well across a wide range of languages. The different datasets we covered in our work include ECHR \cite{chalkidis2019neural} and FSCS \cite{niklaus2021swiss}. Table~\ref{tab:textclassify} outlines these datasets, the prompting methods explored, the cumulative number of languages evaluated, relevant studies, and the most effective prompting strategies.

\footnotesize
\begin{longtable}{p{.6in}p{.9in}p{1.4in}p{.5in}p{.8in}p{.8in}}
    \caption{Prompt Engineering Analysis for Binary Text Classification Task}
    \label{tab:textclassify} \\
    \toprule
    \textbf{Dataset} & \textbf{Prompting Strategy} & \textbf{LLM(s)}  & \textbf{SoTA} & \textbf{\#Languages} & \textbf{References} \\
    \midrule
    \endfirsthead
    
    \multicolumn{6}{c}%
    {{\bfseries \tablename\ \thetable{} -- continued from previous page}} \\
    \toprule
    \textbf{Dataset} & \textbf{Prompting Strategy} & \textbf{LLM(s)}  & \textbf{SoTA} & \textbf{\#Languages} & \textbf{References} \\
    \midrule
    \endhead

    \midrule \multicolumn{6}{c}{{Continued on next page}} \\
    \endfoot
    
    \bottomrule
    \endlastfoot

    ECHR & Native-Basic     & GPT-J-6B, GPT-NeoX-20B     & Native-Basic & 1 & \cite{trautmann2022legal}\\
    \midrule
    FSCS & Native-Basic     & GPT-J-6B, GPT-NeoX-20B     & Native-Basic & 3 & \cite{trautmann2022legal}\\

\end{longtable}

\normalsize
\section{Discussion}



This section presents a comprehensive discussion of the insights drawn from our analysis of multilingual prompt engineering techniques across a wide variety of NLP tasks. Our objective is to explore how core analytical dimensions—including the frequency and type of prompting methods used, the diversity and distribution of NLP tasks, the number of scholarly contributions reviewed and the extent of language coverage—interact with two critical linguistic axes: language families and language resource levels (high-resource vs. low-resource). These axes serve as important structuring devices to understand how existing work is distributed, what disparities persist and where research efforts may need to be rebalanced. By contextualizing the survey findings through these lenses, we aim to offer a more nuanced understanding of the current multilingual prompt engineering landscape and identify patterns that can inform more equitable and effective development of prompting strategies for LLMs.


Language families represent an essential typological classification, encompassing groups of languages that share historical, phonological and grammatical features. Examining how prompting techniques and NLP tasks align with different language families provides important insights into the generalizability and adaptability of these techniques across structurally diverse languages. For instance, many surveyed studies disproportionately focus on Indo-European languages, which often have greater digital representation and institutional support, whereas families such as Niger-Congo, Dravidian or Austroasiatic are far less frequently addressed. This imbalance can lead to methodological biases, where prompting strategies are inadvertently tailored to characteristics common within a single family, thereby limiting cross-family applicability. By analyzing the frequency of prompting methods, the breadth of NLP tasks and the number of cited research works across families, we highlight which linguistic groups are receiving sustained attention and which remain underserved. This form of disaggregated insight is instrumental in promoting linguistic inclusivity and guiding the development of more robust and universally applicable prompting methodologies.


Language resource levels—particularly the distinction between high-resource and low-resource languages—play an equally pivotal role in shaping the landscape of multilingual NLP. High-resource languages benefit from extensive corpora, pretraining data and prior research, enabling more sophisticated prompting techniques and task coverage. In contrast, low-resource languages are often constrained by data scarcity, limited evaluation benchmarks and fewer language-specific LLM adaptations. Understanding how prompting strategies and NLP task diversity vary with resource availability is crucial for evaluating the inclusiveness of current approaches. Our analysis reveals that certain prompting methods, such as CoT variants or translation-based strategies, are more frequently employed in high-resource settings, whereas low-resource languages tend to rely on simpler, direct prompting methods. Additionally, the intersection of resource level and language family provides deeper insight into systemic coverage gaps—such as underrepresentation of low-resource languages from underrepresented families—which may go unnoticed when analyzing each factor in isolation. These findings underscore the importance of developing prompting strategies that are not only performant but also adaptable to linguistic contexts with limited resources, thus contributing to a more globally equitable NLP ecosystem.

\normalsize
\subsection{Language Families}

\begin{figure}
    \centering
        \includegraphics[width=1.0\linewidth]{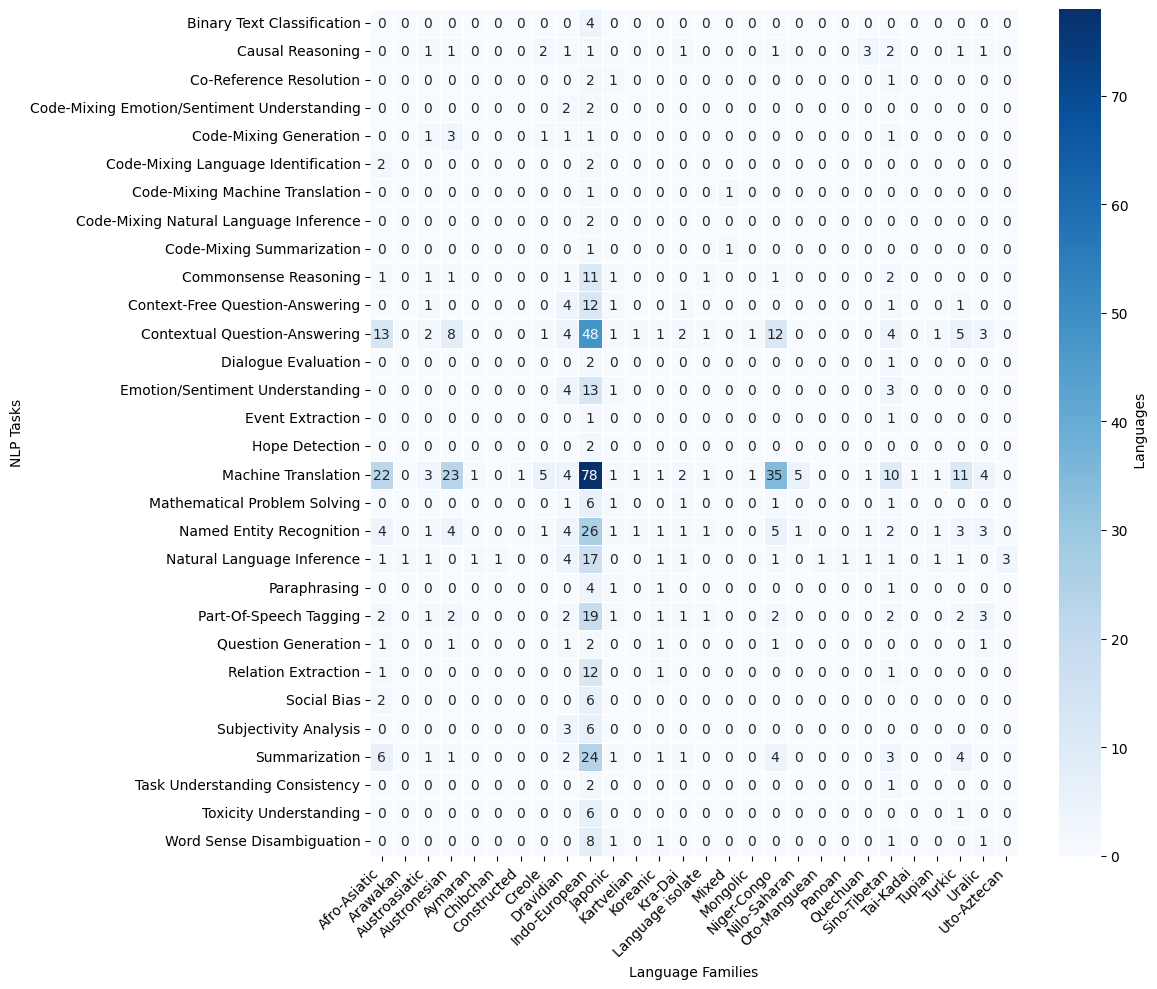}
    \caption{NLP Task Distribution by Language Family}
    \label{fig:nlptaskdislf}
\end{figure}

Examining the intersection of NLP tasks and language families is crucial for understanding both the scope and the inclusivity of current research in multilingual prompt engineering. By analyzing which language families are represented across various NLP tasks, we gain insights into where research efforts are concentrated, which linguistic communities may be underserved and how evenly the benefits of LLMs are distributed globally. The plot in Figure \ref{fig:nlptaskdislf} systematically maps 30 NLP tasks against 28 language families, covering a total of 249 languages covered in this survey and clearly highlights significant disparities in language coverage and task diversity. For instance, the Indo-European family exhibits the broadest coverage, especially dominating tasks such as Machine Translation (with 78 unique languages), Contextual Question-Answering, Summarization, and Named Entity Recognition. This reflects both the prevalence of Indo-European languages in global communication and the accessibility of annotated data. In contrast, language families such as Niger-Congo, Afro-Asiatic and Austronesian are substantially less represented and their coverage is often limited to fewer tasks and languages, frequently centered around Machine Translation. Notably, families such as Austroasiatic, Creole, Quechuan and others appear only sporadically, indicating minimal exploration and a substantial opportunity for future research. The heatmap also reveals that code-mixing tasks such as Code-Mixing Machine Translation or Emotion Understanding are more frequently associated with Dravidian and Indo-European languages, mirroring real-world linguistic environments where multilingual speakers commonly blend languages. Importantly, the underrepresentation of certain language families underscores the need to expand prompt engineering efforts beyond high-resource, widely spoken languages, not only for equitable language technology but also to enhance LLMs’ generalizability and robustness. This granular analysis provides a valuable blueprint for guiding future research towards a more balanced and comprehensive coverage of the world’s linguistic diversity.

\normalsize

\normalsize
\begin{figure}
    \centering
    \includegraphics[width=1.0\linewidth]{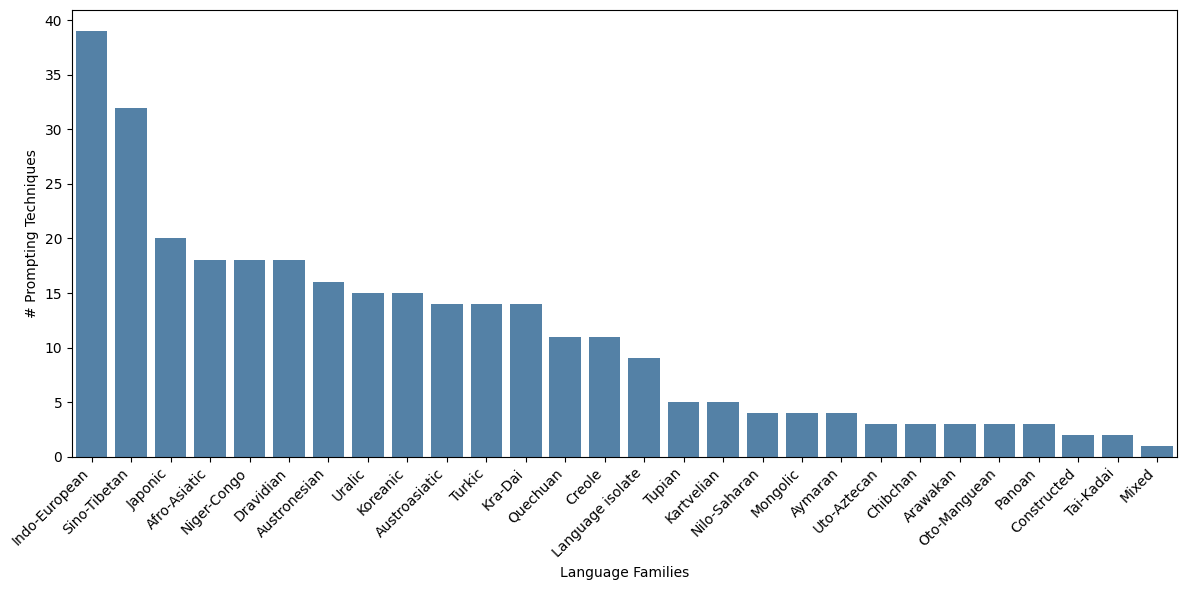}
    \caption{Number of Prompting Techniques by Language Family}
    \label{fig:ptlf}
\end{figure}

Analyzing the distribution of unique prompting techniques across language families is vital for uncovering patterns of methodological innovation and potential gaps in multilingual prompt engineering. Figure~\ref{fig:ptlf} visualizes how 38 distinct prompting techniques are distributed over 28 language families. Such analysis helps illuminate not only where research creativity and methodological diversity are flourishing but also which linguistic communities might benefit the most or the least from current advancements in prompt engineering. As shown in the plot, the Indo-European language family stands out with the highest number of unique prompting techniques, closely followed by Sino-Tibetan and Japonic families. This dominance can be attributed to the extensive availability of data, broader international research focus and the central role these languages play in both global communication and NLP research. Families such as Afro-Asiatic, Niger-Congo, Dravidian and Austronesian also exhibit moderate methodological diversity, indicating a healthy, albeit more focused, experimentation with prompting strategies. In contrast, the right side of the plot reveals a significant drop-off, with language families like Chibchan, Oto-Manguean and Tai-Kadai, as well as Constructed and Mixed language groups, being associated with very few prompting techniques. This skewed distribution signals that much of the innovation in multilingual prompt engineering is concentrated around a handful of high-resource language families, while low-resource and geographically isolated families remain relatively neglected. This underlines a critical opportunity for future research: expanding the methodological toolkit to better serve underrepresented languages and families. Such expansion is not only essential for the ethical and inclusive development of language technology but is also likely to drive advancements in model robustness, adaptability and cross-linguistic generalization. By systematically mapping these methodological disparities, this analysis provides a foundation for setting future research priorities and fostering a more equitable landscape in multilingual NLP.

\normalsize
\begin{figure}
    \centering
    \includegraphics[width=1.0\linewidth]{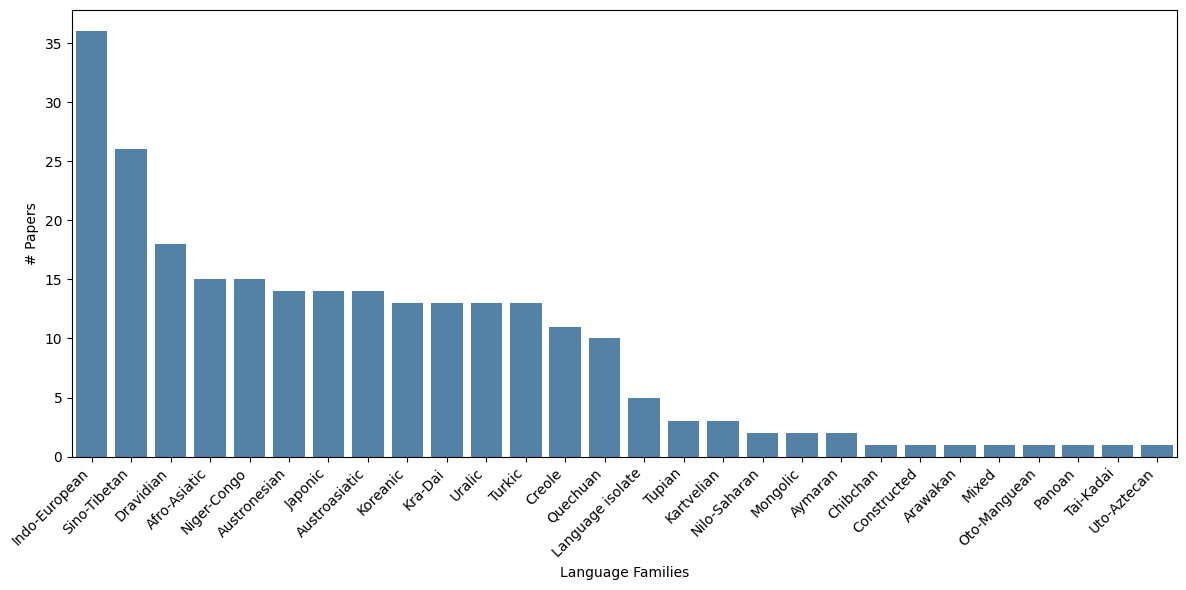}
    \caption{Number of Papers by Language Family}
    \label{fig:nplf}
\end{figure}

Studying the distribution of research papers across language families provides a clear perspective on where the focus of the multilingual prompt engineering community currently lies and highlights gaps that require further scholarly attention. Figure~\ref{fig:nplf} visualizes the number of unique research papers, out of a total of 36 covered in this survey, that have investigated prompt engineering within each of the 28 identified language families, collectively representing 249 languages. This analysis is essential for understanding not just the breadth of multilingual coverage, but also the research intensity and scholarly engagement with different linguistic communities. The figure shows a pronounced concentration of research activity around Indo-European and Sino-Tibetan languages, with the Indo-European family leading by a substantial margin—over 35 papers address this family, followed by more than 25 for Sino-Tibetan. This prominence is unsurprising given the global dominance and resource availability for these languages, which drive both academic interest and practical NLP applications. The Dravidian, Afro-Asiatic, Niger-Congo and Austronesian families also demonstrate moderate research engagement, each with around 15-19 papers, suggesting a growing but still limited recognition of the importance of linguistic diversity in NLP research. In contrast, families such as Tupian, Kartvelian, Nilo-Saharan, Mongolic and others have very few papers, typically 1 to 5 underscoring a substantial underrepresentation. Even less attention is given to smaller or more geographically isolated families like Chibchan, Oto-Manguean and Uto-Aztecan, as well as Constructed or Mixed languages, where only 1 or 2 papers have addressed prompt engineering. This pronounced skew highlights the need for broader and more inclusive research efforts to ensure that advancements in multilingual prompt engineering benefit a wider spectrum of the world’s linguistic landscape. The analysis also suggests that future work should prioritize low-resource and underrepresented language families, both to foster equitable access to language technologies and to enhance the robustness and adaptability of LLMs in genuinely multilingual contexts.

\normalsize
\begin{figure}
    \centering
    \includegraphics[width=1.0\linewidth]{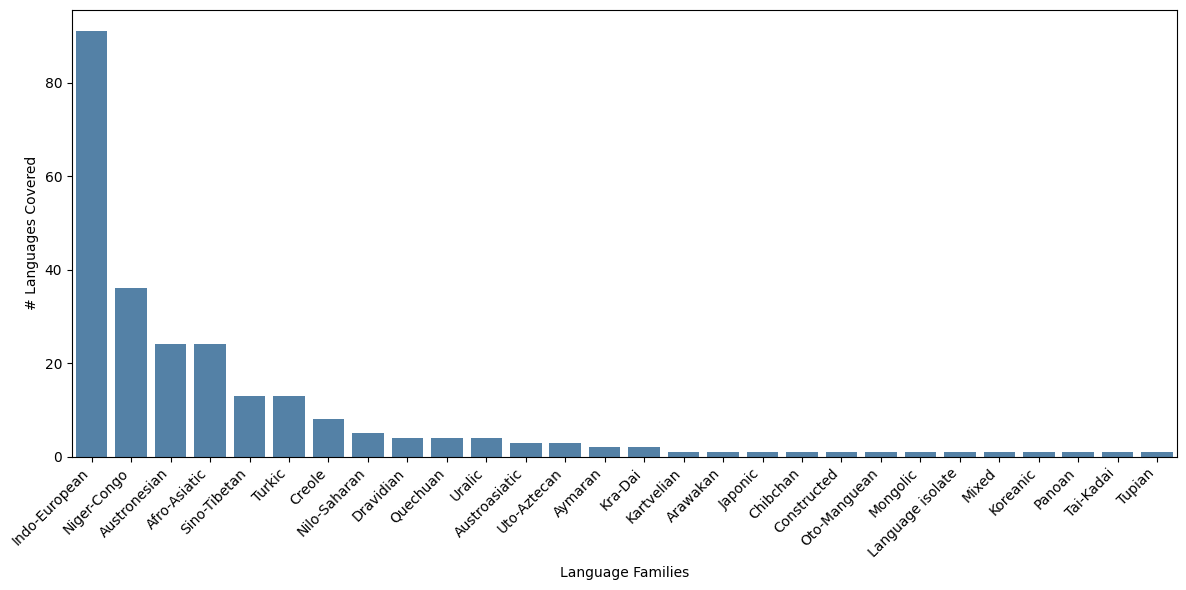}
    \caption{Language Coverage by Language Family}
    \label{fig:langcovlf}
\end{figure}

Investigating the distribution of languages across language families is essential for understanding the true breadth and representation of multilingual prompt engineering research. This type of analysis allows us to gauge not only which linguistic groups receive the most attention but also where significant representation gaps remain. The bar plot in Figure~\ref{fig:langcovlf} showcases the number of languages covered within each of the 28 language families represented in our survey, which collectively encompasses 249 languages. The Indo-European family stands out as the most extensively represented, with nearly 90 languages included—a reflection of both its global prevalence and its prominence in NLP research. Following this, the Niger-Congo, Austronesian and Afro-Asiatic families exhibit moderate coverage, each accounting for more than 20 languages. This indicates a growing but still limited effort to extend prompt engineering research to linguistically diverse and populous regions of the world, such as Sub-Saharan Africa and Southeast Asia. In contrast, several language families remain underrepresented, with only a handful of languages included in each. Families such as Uto-Aztecan, Aymaran, Chibchan, Oto-Manguean and Tai-Kadai appear at the far right of the plot, highlighting the scarcity of research and data for these languages. Some families—including Mongolic, Language Isolate, Mixed, and Koreanic are represented by just 1 or 2 languages, suggesting that NLP advances for these groups remain limited to a small subset of the linguistic landscape. Notably, the Creole family also features relatively few languages despite its sociolinguistic importance in several world regions. The observed disparities highlight the necessity for more targeted research and resource development for low-resource and underrepresented language families, ensuring that future advances in prompt engineering can better serve the world’s linguistic diversity. By systematically identifying these gaps, this analysis provides a roadmap for expanding the scope and equity of multilingual LLM research going forward.

\normalsize
\subsection{High-Resource Vs Low-Resource}

\begin{figure}
    \centering
    \includegraphics[width=1.0\linewidth]{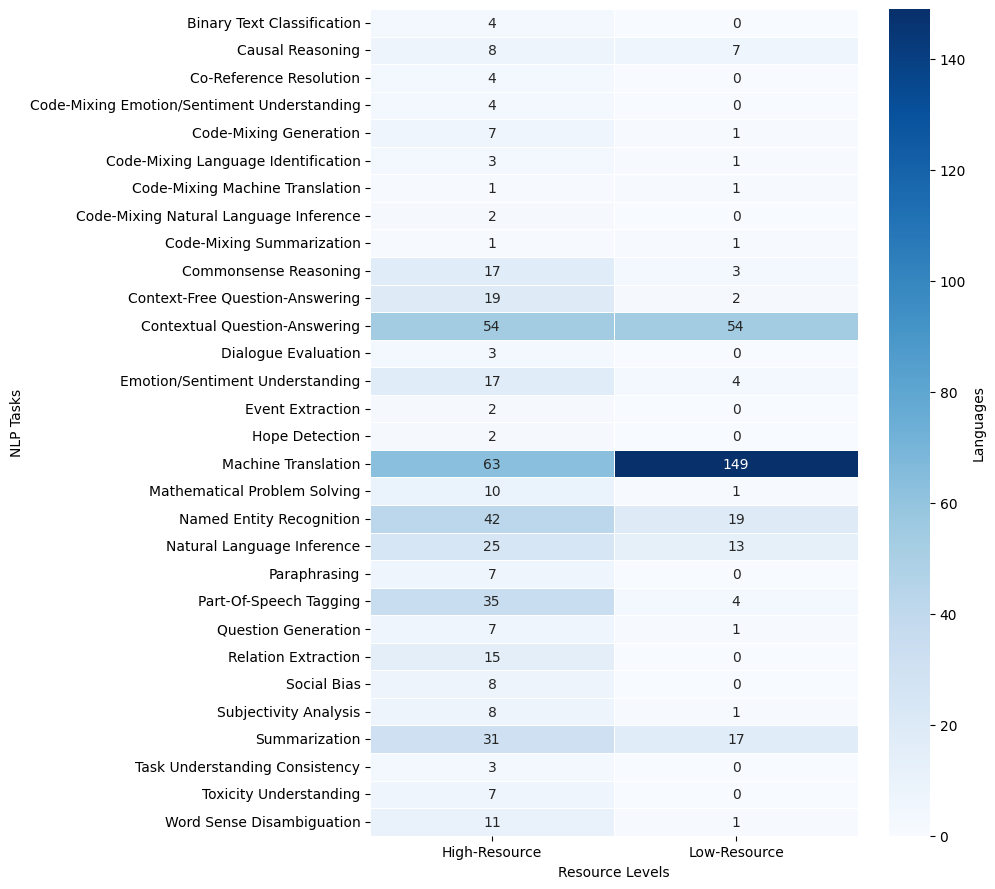}
    \caption{NLP Task Distribution by Resource Level}
    \label{fig:nlptaskdishrlr}
\end{figure}

Studying the distribution of NLP tasks across language resource levels is significant for assessing the inclusivity and equity of multilingual prompt engineering research. High-resource and low-resource languages represent fundamentally different challenges for LLMs: while high-resource languages benefit from abundant data and research attention, low-resource languages often suffer from limited datasets and less robust model performance. The heatmap in Figure~\ref{fig:nlptaskdishrlr} illustrates the coverage of 30 NLP tasks with respect to these two resource levels, spanning 249 languages in total. Notably, a striking imbalance emerges as most NLP tasks are heavily concentrated in high-resource languages, as evidenced by the consistently higher counts in the left column for tasks such as Named Entity Recognition (42 languages), Context-Free Question-Answering (54) and Summarization (31). In contrast, their low-resource counterparts typically see far fewer languages addressed, highlighting the disparity in research focus and resource allocation. Machine Translation stands out as a key exception: it demonstrates the broadest coverage among all tasks, engaging 63 high-resource languages and an impressive 149 low-resource languages. However, outside of translation, tasks such as Mathematical Problem Solving, Word Sense Disambiguation and Subjectivity Analysis remain almost exclusively confined to high-resource languages, with negligible or no coverage in low-resource settings. The heatmap thus makes it evident that while efforts are growing to extend multilingual LLM research, there is still a pressing need for targeted advancements in prompt engineering and resource creation for the world’s less represented languages, especially beyond the translation task. Moving forward, it is crucial that the community prioritizes the development and evaluation of prompting strategies that are effective for low-resource settings, thereby advancing language technology for a broader and more inclusive set of users worldwide.

\normalsize

\normalsize
\begin{figure}
    \centering
    \includegraphics[width=0.3\linewidth]{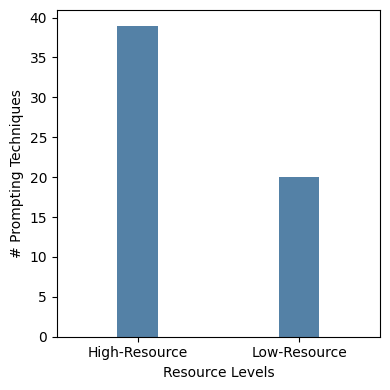}
    \caption{Number of Prompting Techniques by Resource Level}
    \label{fig:ptrl}
\end{figure}

Understanding the distribution of prompting techniques across language resource levels is essential for assessing how innovation in prompt engineering reaches both well-resourced and under-resourced languages. The bar chart in Figure~\ref{fig:ptrl} provides a clear comparison, plotting the number of unique prompting techniques applied to high-resource and low-resource languages. The results reveal a significant disparity: all 38 distinct prompting strategies have been deployed in high-resource language contexts, while only 20 approaches have been extended to low-resource languages. This difference highlights the persistent gap in research attention and methodological experimentation between these two groups. High-resource languages benefit from a greater diversity of prompting approaches, which is likely driven by the availability of large-scale annotated datasets and the prioritization of these languages in LLM development and evaluation pipelines. The breadth of experimentation in high-resource settings contributes to rapid advancements in performance, robustness and nuanced capabilities for different NLP tasks. In contrast, low-resource languages see far less methodological variety, limiting the potential for discovering effective strategies that might address their unique linguistic challenges.

\normalsize
\begin{figure}
    \centering
    \includegraphics[width=0.3\linewidth]{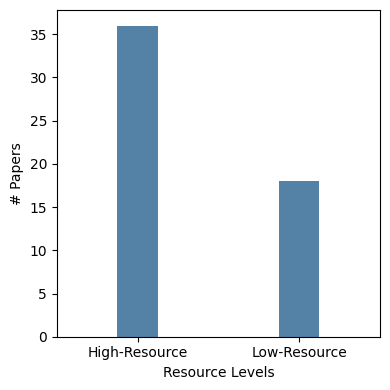}
    \caption{Number of Papers by Resource Level}
    \label{fig:nplr}
\end{figure}

Examining the distribution of research papers focused on high-resource versus low-resource languages is instrumental in understanding the landscape of scholarly attention within multilingual prompt engineering. The bar chart in Figure~\ref{fig:nplr} visualizes this divide by comparing the number of unique research papers that investigate prompting techniques in high-resource and low-resource language contexts, across the 249 languages and 36 papers covered in this survey. The findings reveal a notable imbalance: research into high-resource languages outpaces that on low-resource languages nearly two to one, with 36 papers addressing the former and just 18 the latter. This discrepancy underscores the persistent tendency of the NLP research community to prioritize languages for which extensive data, benchmarks and pre-trained models are readily available. High-resource languages benefit from a richer ecosystem of experimentation, methodological diversity and cross-pollination between research groups, resulting in faster progress and broader impact for these languages in practical applications. On the other hand, the comparatively sparse body of work dedicated to low-resource languages reflects ongoing challenges—such as data scarcity, lack of standardized evaluation resources and reduced accessibility for non-major languages.

\begin{figure}
    \centering
    \includegraphics[width=0.3\linewidth]{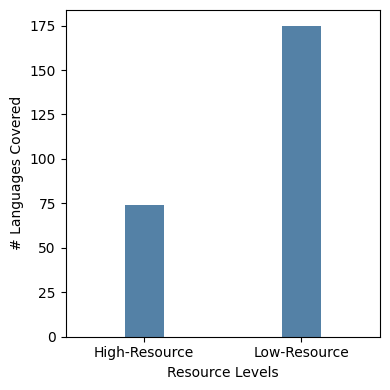}
    \caption{Language Coverage by Resource Levels}
    \label{fig:langcovlr}
\end{figure}

Delving into the coverage of languages across different resource levels is pivotal for evaluating the global reach of multilingual prompt engineering research. The bar chart in Figure~\ref{fig:langcovlr} provides a comparative overview of the number of languages from high-resource and low-resource categories that have been explored in the survey, spanning a total of 249 languages. Notably, while the majority of research papers and prompting techniques focus on high-resource languages (as shown in previous plots), this figure highlights a remarkable breadth in low-resource language inclusion—showing that low-resource languages constitute a far greater proportion of the overall language coverage, with over 175 languages represented compared to approximately 74 high-resource languages. This contrast reveals both the challenges and the emerging opportunities in the field. High-resource languages are typically favored due to the availability of extensive corpora, benchmarks and robust toolchains, resulting in a dense concentration of research activity. However, the significant number of low-resource languages featured in prompting-based studies suggests growing efforts to bridge the digital divide. Such broad representation can be attributed to advances in cross-lingual transfer learning, the use of massively multilingual datasets and collaborative initiatives targeting underserved linguistic communities. Despite fewer papers and prompting methods dedicated to low-resource settings, the wide linguistic coverage demonstrates that prompt engineering research is beginning to extend its reach and potentially democratize language technology.

\normalsize

\normalsize
\section{Conclusion}

Multilingual prompt engineering has emerged as a critical technique for enhancing the performance of LLMs across diverse linguistic landscapes. In this survey, we conduct a comprehensive review of 36 research papers, covering 39 multilingual prompting techniques applied to 30 NLP tasks. We present a taxonomy diagram of multilingual prompting techniques and independently propose a standardized categorization of these methods across various NLP tasks. Furthermore, we identify potential SoTA prompting strategies for a range of multilingual NLP datasets and analyze trends in prompting usage across different language families. This includes examining the most frequently explored prompting methods by language family and resource type, offering a structured overview of how prompting techniques are applied across diverse linguistic contexts.

Our survey highlights the growing role of multilingual prompt engineering in optimizing LLM performance across diverse linguistic settings. Although significant progress has been made, challenges such as cross-lingual consistency, dataset limitations and the standardization of evaluation metrics persist. By consolidating existing approaches and identifying trends, this survey provides a foundation for future advancements in multilingual prompt engineering and its applications in global NLP. As research in this area continues to evolve, further studies can explore the effectiveness of different prompting strategies across underrepresented languages and task-specific contexts, ultimately contributing to more effective and accessible multilingual language models.

\bibliography{iclr2024_conference}
\bibliographystyle{iclr2024_conference}


\end{document}